\documentclass[10pt,journal,compsoc]{IEEEtran}

\usepackage{amsmath,amssymb,amsfonts}

\usepackage[dvipsnames,table]{xcolor}
\definecolor{myblue}{rgb}{.894,.937,.965}
\definecolor{mydark_blue}{rgb}{.68,.776,.906}
\definecolor{NO3}{rgb}{.749,.902,.808}
\definecolor{NO2}{rgb}{.999,.973,.773}
\definecolor{NO1}{rgb}{0.98, 0.78, 0.57}

\usepackage{graphicx}

\usepackage{array}
\usepackage{booktabs}
\usepackage{multirow}
\usepackage{multicol}
\usepackage{colortbl}
\usepackage{makecell}   
\usepackage{adjustbox}  

\usepackage[caption=false,font=normalsize,labelfont=sf,textfont=sf]{subfig}
\usepackage{stfloats}
\usepackage{rotating}

\usepackage{algorithm}
\usepackage{algorithmic}

\usepackage{textcomp}
\usepackage{url}
\usepackage{verbatim}
\usepackage{cite}
\usepackage{soul}

\usepackage[breaklinks,colorlinks]{hyperref}

\usepackage{pifont}
\newcommand{\cmark}{\ding{51}}   
\newcommand{\xmark}{\ding{55}}   

\renewcommand{\rotcell}[1]{\parbox{2cm}{\centering #1}}

\ifCLASSINFOpdf
\else
\fi

\begin{document}

\title{Towards Interactive Video World Modeling: Frontiers, Challenges, Benchmarks,\\ and Future Trends}

\author{Jiuming~Liu,
       Chaojun~Ni, Mengmeng~Liu, Chensheng~Peng, Fangjinhua~Wang, Sitian~Shen,\\ Marc~Pollefeys,~\IEEEmembership{Fellow,~IEEE}, Masayoshi Tomizuka,~\IEEEmembership{Life~Fellow,~IEEE}, Ayush Tewari, \\and Per Ola~Kristensson
\IEEEcompsocitemizethanks{
\IEEEcompsocthanksitem Jiuming Liu, Ayush Tewari, and Per Ola Kristensson are with the Department of Engineering, University of Cambridge, U.K. (email: liujiuming123@gmail.com, at2164@cam.ac.uk, pok21@cam.ac.uk) \protect
\IEEEcompsocthanksitem Chaojun Ni is with the Peking University, China. \protect
\IEEEcompsocthanksitem Mengmeng Liu is with the University of Twente, Netherlands. \protect
\IEEEcompsocthanksitem Chensheng Peng and Masayoshi Tomizuka are with the Mechanical Systems Control Laboratory, University of California, Berkeley, USA. \protect
\IEEEcompsocthanksitem Fangjinhua Wang and Marc Pollefeys are with the ETH Zurich, Zurich, Switzerland. Marc Pollefeys is additionally with Microsoft. \protect
\IEEEcompsocthanksitem Sitian Shen is with the University of Oxford, UK.
 }

}

\markboth{Journal of \LaTeX\ Class Files,~Vol.~14, No.~8, May~2026}%
{Liu \MakeLowercase{\textit{et al.}}: Towards Interactive Video World Modeling: Frontiers, Challenges, Benchmarks, and Future Trends}


\IEEEtitleabstractindextext{
\begin{abstract}
With rapid development of large language models and diffusion-based content generation, world modeling has attracted increasing research attention, benefiting various downstream domains such as game engines, embodied AI, autonomous driving, etc. Through explicitly incorporating user actions into world state transition, recent literature empowers world modeling with interactivity in an action-conditioned video or 3D generation paradigm, further enhancing controllability over world evolutions and facilitating users to freely traverse, manipulate, navigate, and personalize the state evolution. In this paper, we aim to systematically review recent research trends, technical developments, evaluation benchmarks, and also propose future potential directions in interactive world modeling. Specifically, we first summarize recent efforts and trends in terms of application scenarios, world state evolution, and scene modality. Afterwards, we delve into three crucial technical challenges, including action-conditioned controllability, long-horizon interactions and memory, and action-following responsiveness for real-time interactivity. Furthermore, we also thoroughly compare existing benchmarks and metrics in four specific application fields: open-world exploration, game engine, autonomous driving, and robotics. Finally, we discuss several promising future directions in achieving next-generation interactive world modeling. The corresponding repository is publicly available at: \url{https://github.com/liujiuming123/Awesome-Interactive-World-Model}.
\end{abstract}

\begin{IEEEkeywords}
World model, Action-conditioned video generation, Interactive user interface, Long-horizon consistency, Memory mechanism, Real-time interactivity, Open-world exploration, Immersive game engine.
\end{IEEEkeywords}}

\maketitle

\IEEEdisplaynontitleabstractindextext

\IEEEpeerreviewmaketitle

\IEEEraisesectionheading{\section{Introduction}}
\emph{``Tell me and I forget, teach me and I remember, involve me and I learn.''} --- Benjamin Franklin \\

\IEEEPARstart{W}ith rapid development of Artificial Intelligence Generated Content (AIGC) and multi-modal large language models (LLMs), world models have gained increasing research focus \cite{craik1967nature,ha2018world,lecun2022path}. By formulating the underlying dynamics of real-world environments, world models can provide fundamental understanding and generate counterfactual future predictions based on historical observations and actions, widely facilitating domains in autonomous driving \cite{wu2024ivideogpt,wang2024driving,wang2024drivedreamer}, embodied AI \cite{li2025comprehensive,zhong2025unrealzoo,bar2025navigation,lv2026viva}, and game engines \cite{zhang2025matrix,he2025matrix,Yu_2025_ICCV}. 

\textit{Interactivity} constitutes a fundamental property of world models. Humans perceive and comprehend the real world by consistently interacting with the environment, thereby progressively improving their cognitive and reasoning ability. The same applies to machine intelligence \cite{lake2017building}. Recently, higher-level interactivity \cite{bruce2024genie,he2025matrix} with frame- or region-based interventions has attracted increasing attention in world modeling, allowing users to intervene in world evolution and immersively explore, manipulate, navigate, reason, and plan within generated environments. Interactive world models \cite{parkerholder2024genie2,genie3,bar2025navigation,zhang2025matrix,he2025matrix,Yu_2025_ICCV} have been widely explored in both industry and academia as in Table \ref{tab:survey1}. As a milestone, Genie series \cite{parkerholder2024genie2,genie3,bruce2024genie} from Google have pioneered a new industrial frontier for general interactive world models, which can predict minute-level videos supporting real-time navigation up to 24 fps. WorldGen \cite{wang2025worldgen} launched by Meta designs a text-guided traversable 3D world model based on instance-aware scene decomposition and procedural scene generation. NVIDIA proposes Lyra 2.0 \cite{shen2026lyra2} for explorable 3D-consistent generation deployable in simulation engines. There are also many industry-level products for general interactive world models, such as PixelVerse R1 \cite{pixelverse}, Happy Oyster \cite{happy}, GWM-1 \cite{runway}, etc. In terms of open-world exploration, recent Wonder-series \cite{yu2024wonderjourney,yu2025wonderworld,cao2025wonderzoom} empower people to traverse from anywhere to everywhere given a reference image. LingBot-World \cite{team2026advancing} renders physically plausible dynamic scenes. As for game engines, several successful attempts have emerged, including Matrix-Game \cite{zhang2025matrix,he2025matrix,matrixgame3}, Yume \cite{mao2025yume1,mao2025yume1.5}, HY-WorldPlay \cite{sun2025worldplay}, etc. Despite tremendous advancement, interactive world modeling still faces a series of challenges, such as long-horizon consistency during interaction, real-time responsiveness to user actions, effective and generalizable action injection manners, efficient memory retrieval, etc \cite{li2025vmem,savva2026solaris,liu2026realwonder}. However, we observe that the current literature lacks a comprehensive survey of recent advances, unsolved challenges, commonly-adopted benchmarks, and potential future research trends in this field.

\begin{figure*}[t]
\centering
\includegraphics[width=1.0\linewidth]{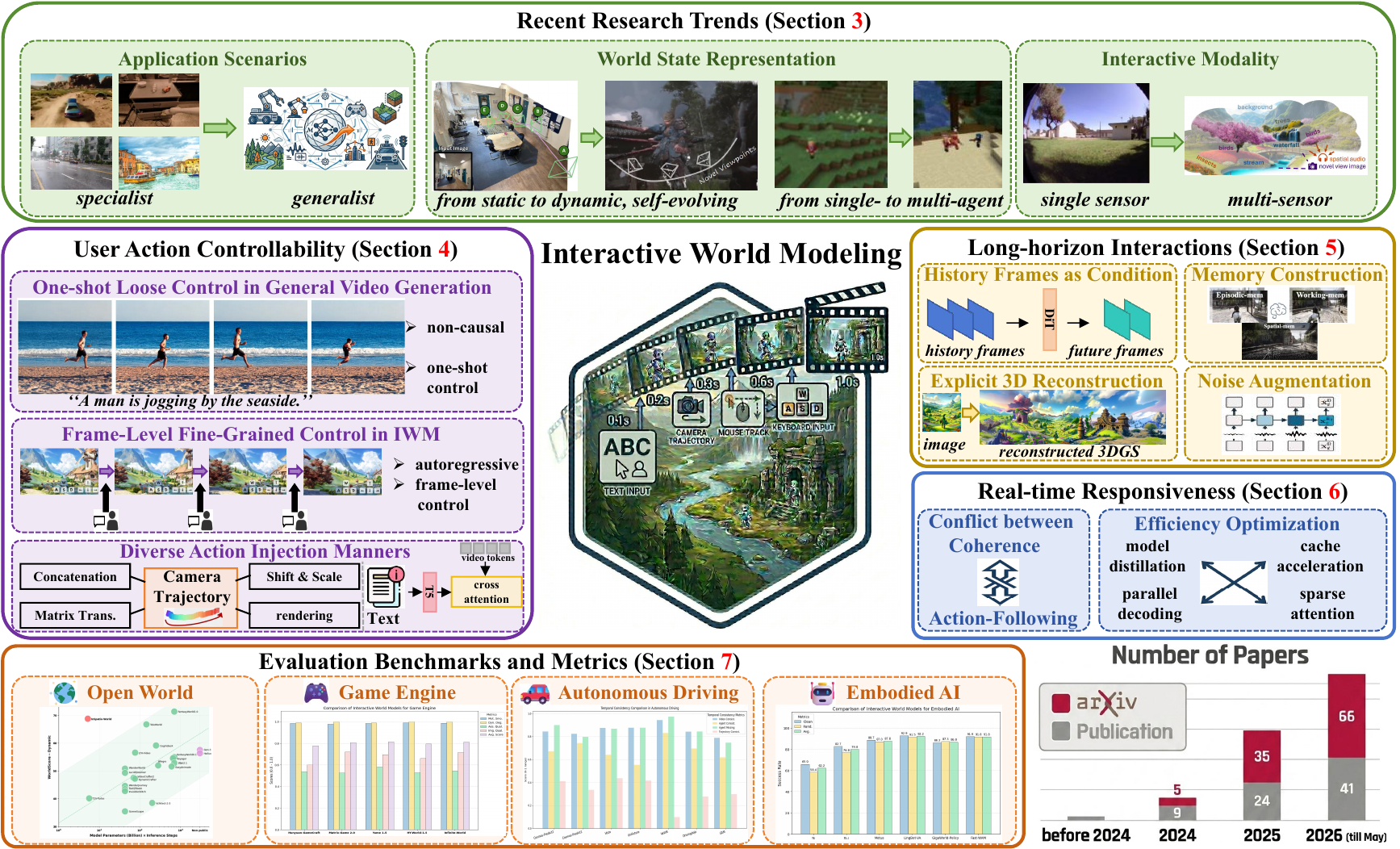}
\vspace{-7mm}
\caption{\textbf{Overview of our survey structure.} The survey begins by thoroughly reviewing recent research trends, including application scenarios, world states, interactive modalities. Then, three key technique bottlenecks are systematically extended, ranging from user controllability from actions, long-horizon interactions, and real-time responsiveness. Finally, we list existing evaluation benchmarks and compare metrics across four applications: embodied AI, autonomous driving, game engines, and open-world exploration. We also calculate the evolving paper numbers until now. }
\label{fig:pipiline}
\vspace{-3mm}
\end{figure*}

To address this gap, in this survey paper, we aim to comprehensively review recent efforts and progress in interactive world modeling, highlighting key research directions and unresolved challenges toward next-generation interactive world models. As illustrated in Fig. \ref{fig:pipiline}, we first introduce some background information in Section \ref{sec:back}, including specific definitions, diverse interaction interfaces, and advanced generation backbones. Afterwards, we summarize recent research trends in Section \ref{sec:trend}: from specific to general application scenarios; from static, single-agent to dynamic, self-evolving, multi-agent world states; and from single sensory to multi-sensory interacting modalities. Furthermore, we revisit three key technical bottlenecks in the interactive world modeling domain: action-conditioned controllability (Section \ref{sec:control}), long-horizon interactions and memory mechanism (Section \ref{sec:long-term}), and real-time action responsiveness (Section \ref{sec:response}). In Section \ref{sec:evaluate}, we summarize commonly-used evaluation benchmarks, datasets, and metrics in interactive world modeling, which will facilitate a global understanding of what advances this field has achieved and comparisons among various latest methods across embodied AI, autonomous driving, game engine, and open-world exploration. Finally, unsolved challenges and promising future research directions are analyzed in Section \ref{sec:future}. We hope this survey provides a comprehensive overview of the field and facilitates future research in interactive world model.


\section{Preliminaries}
\label{sec:back}
\subsection{Definition of Interactive World Model}
\label{sec:definition}
\subsubsection{World Model} 
\textbf{Evolution History.} The concept can be traceable to the mid-twentieth century, when a \textit{mental representation model} was proposed by Cambridge psychologist, philosopher, and control theorist Kenneth Craik in his book `The Nature of Explanation'. This model is assumed to mirror the structural properties of the external world and can act as an automatic regulator from philosophical and cybernetic views \cite{craik1967nature}. Furthermore, this mental model should also possess flexible prediction capability, which is further described as \textit{counterfactual competence}, capable of modeling the causal nature of the natural world and generating future predictions after a series of interventions \cite{danwilliam}. In cognitive science, some researchers aim to build human-like learning and thinking machines with world causality modeling, intuitive learning of physics and psychology, and generalization ability to new tasks \cite{lake2017building,wong2023word}. In recent years, the concept of \textit{world model} \cite{ha2018world,lecun2022path} has been systematically mentioned to learn the internal environmental dynamics by predicting future states from historical observations and actions, enabling an intelligent agent to imagine, reason, navigate, and make policies. 

\noindent\textbf{Mathematical Definition.} The Partially Observable Markov Decision Process (POMDP) with tuple $(\mathcal{S}, \mathcal{A}, \mathcal{O}, \mathcal{T}, \mathcal{R})$ is commonly adopted to represent world models, where $\mathcal{S}$ indicates the world state, $\mathcal{A}$ is a set of actions carried out by humans or agents, and $\mathcal{O}$ is a set of visual observations. The transition function $\mathcal{T}$ can describe internal world evolutions as $p(\mathrm{s}_{t+1}|\mathrm{s}_{t},\mathrm{a}_{t})$. A policy can learn to choose actions that lead to high rewards $\mathcal{R}$ extracted from generated videos. According to \cite{ha2018world}, world models can both abstract world states $\mathcal{S}$ and illustrate world evolutions $\mathcal{T}$.

\noindent\textbf{Diverse Exploration Paths.} Currently, there are different paths in achieving effective world modeling. Some methods \cite{yang2026neoverse,he2025matrix,wang2025worldgen} adopt a pixel-level prediction based on pre-trained video backbones, e.g., Wan 2.2 \cite{wan2025wan}, learning generalizable priors from large-scale internet videos. There are also other parallel venues, such as the Joint Embedding Predictive Architecture (JEPA) proposed by \cite{lecun2022path} that leverages latent space prediction rather than per-pixel prediction. Some researchers also emphasize spatial intelligence with explicit 3D or 4D world representations such as Marble \cite{marble}, PhysGen3D \cite{chen2025physgen3d}, VDAWorld \cite{o2025vdaworld}, etc. 

Furthermore, we notice that early world models commonly suffer from limited interactivity. To address this problem, increasing recent works emphasize the interactivity by transforming passive video generation into active intervention with users' progressive actions. In this survey, we mainly focus on \textbf{video-based interactive world models}, which enable users or agents to exert fine-grained control actions, e.g., keyboard input, and have more immersive experience, such as navigating within the generated world.



\begin{figure}[t]
\centering
\includegraphics[width=1.0\linewidth]{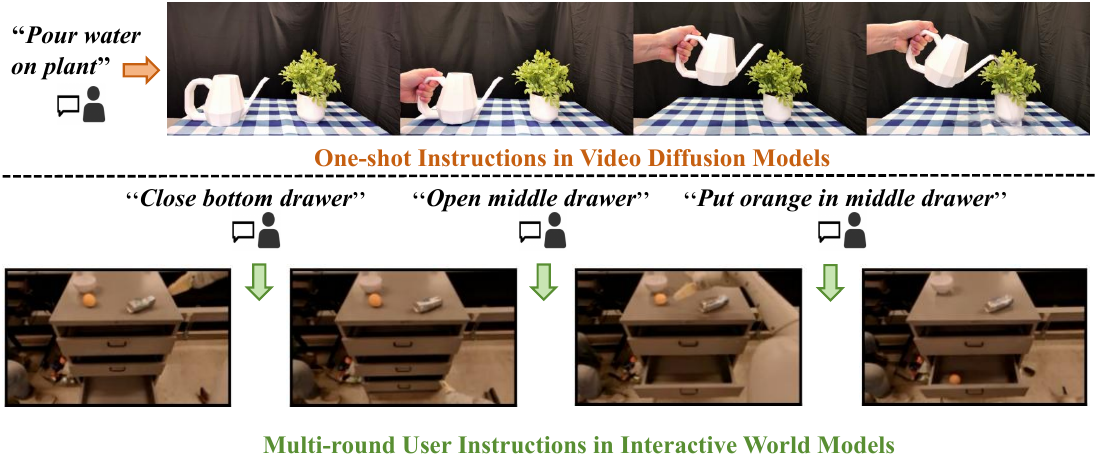}
\vspace{-6mm}
\caption{\textbf{Comparison of action controllability between video diffusion models and interactive world models.} Previous video diffusion models only exert \textit{one-shot} instructions \cite{singer2022make}, while interactive world models leverage \textit{frame-level multi-round} instructions \cite{yang2024learning}.}
\label{fig:compare_control}
\vspace{-3mm}
\end{figure}

\subsubsection{Interactive Video World Models} 
Unlike conventional video generation models that synthesize a complete video from a one-shot condition, an interactive world model emphasizes closed-loop generation, where users or agents can continuously intervene in the generated world in Fig. \ref{fig:compare_control}. Basically, an interactive world model recurrently generates future observations by incorporating the interaction history and newly issued controls \cite{yang2024learning}:
\begin{equation}
\mathbf{o}_{t+1} \sim p_{\phi}
\left(
\mathbf{o}_{t+1}
\mid
\mathcal{H}_{t}, \mathbf{a}_{t}, \mathbf{c}_{t}
\right),
\label{eq:interactive_world_model}
\end{equation}
where $\mathbf{o}_{t+1}$ is the next observation. $p_{\phi}$ is the interactive generative world model parameterized by $\phi$. $\mathcal{H}_{t}$ is the interaction history. $\mathbf{a}_{t}$ is the current user or agent action. And $\mathbf{c}_{t}$ is the additional condition in the time step $t$, such as text, image, or editing control as a reference. Specifically, the interaction history is defined as:
\begin{equation}
\mathcal{H}_{t} =
\left\{
\mathbf{o}_{1:t},
\mathbf{a}_{1:t-1},
\mathbf{c}_{1:t-1}
\right\},
\label{eq:interaction_history}
\end{equation}
where $\mathbf{o}_{1:t}$, $\mathbf{a}_{1:t-1}$, and $\mathbf{c}_{1:t-1}$ denote previous observations, actions, and conditions, respectively. This formulation distinguishes interactive world modeling from passive video generation by requiring progressive updates according to multi-round user interventions.



In summary, an interactive world model should exhibit the following characteristics:
\begin{itemize}
  \item \textbf{Multi-round and Fine-grained Controllability by Users (Human-in-the-Loop).} Unlike prior video generation methods which only implement a \textit{one-shot} instruction \cite{zhang2023adding}, interactive world models require that the generation process is controlled by more fine-grained user instructions at the \textit{frame-level} \cite{li2025magicworld} or even \textit{region-level} \cite{wang2026worldcompass} as in Fig. \ref{fig:compare_control}. After each interaction round, users progressively execute another instruction based on the latest output.

  \item \textbf{Long-horizon and Consistent World Transition.} In response to user instructions, an interactive world model can continuously update the world state, generate long-duration visual content, and preserve consistent scene transitions across multiple rounds of user interaction \cite{li2025magicworld,liao2025genie}. 

  \item \textbf{Real-time Interaction and Feedback.} Another characteristic of interactive world models is the immediate online feedback. Distinct from offline video generation models, interactivity naturally imposes a strong constraint on response efficiency, offering users a more immersive experience with instant visual feedback.
\end{itemize}

In the main chapters, we will revolve around these three attributes above: `Action Controllability' (Section \ref{sec:control}), `Long-Horizon Consistency and Memory' (Section \ref{sec:long-term}), and `Real-time Responsiveness' (Section \ref{sec:response}).

\subsection{Diverse Interaction Interfaces}
Interfaces serve as a \textit{bridge} between humans and world models, enabling users to execute controls for exploration, manipulation, and interaction within generated worlds \cite{bruce2024genie}. Overall, concurrent interfaces mainly consist of:

\textbf{(1) Visual Interface}: One can customize the generated contents by providing reference images \cite{yu2025wonderworld}, videos \cite{wu2024ivideogpt}, layouts \cite{wang2025worldgen}, dragging \cite{wang2025world}, or hand-written sketches \cite{bruce2024genie}. Recent game engines \cite{zhang2025matrix,he2025matrix,mao2025yume1} also incorporate keyboard or mouse input for immersive gameplay experience.


\textbf{(2) Text Interface}: Compared to visual reference, text instructions provide greater flexibility in specifying user intentions. By leveraging advanced large-scale pre-trained language models, such as T5 \cite{raffel2020exploring}, users can modify the style, weather, lighting conditions, or add/delete specific objects. 


\textbf{(3) Audio Interface}: Increasing efforts focus on introducing audio guidance when exploring and interacting with the generation models. For example, Pixelverse-R1 \cite{pixelverse} launches their next-generation world model providing audio commands, where users can personalize the future event or change the generated environments through speech.

\textbf{(4) Physical Interface}: Obeying the physical principle is another significant property of world models. Recently, some researchers explicitly embed physical 3D point force \cite{liu2026realwonder}, gravity field \cite{zhan2026perpetualwonder}, etc, to faithfully imitate real-world dynamics and strengthen spatial intelligence.

\textbf{(5) Other Interfaces}: Apart from the above widely-used interfaces, there are also various interfaces where people can conduct interactive instructions. For example, IWS \cite{wang2026interactive} uses teleoperation as an interface on which users can leverage robotic commands.

\begin{figure}[t]
\centering
\includegraphics[width=1.0\linewidth]{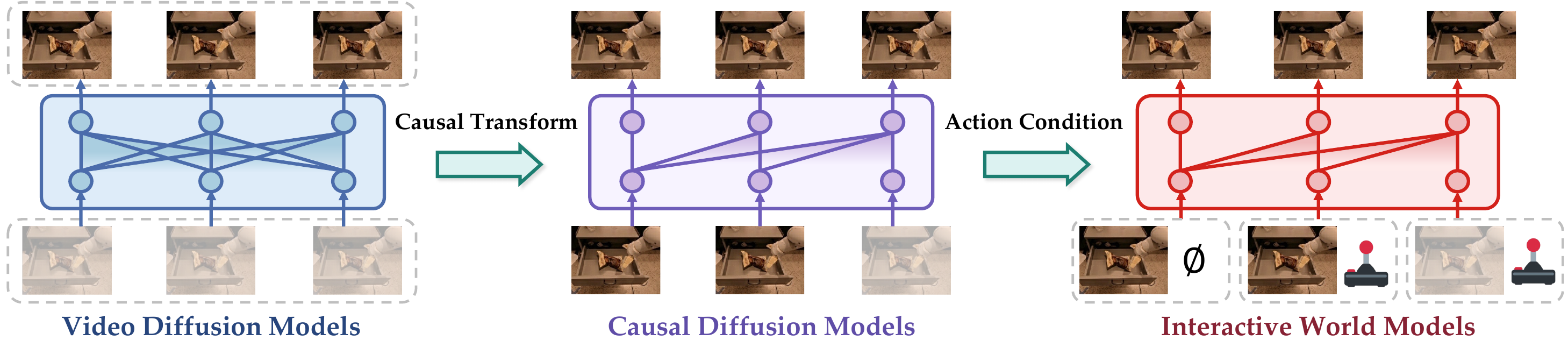}
\vspace{-6mm}
\caption{\textbf{Transformation from video diffusion models to interactive world models.} Unlike prior general video diffusion models which simultaneously output all video frames with bi-directional temporal cues, transforming them into interactive world models requires both causality establishment and action condition \cite{huang2026vidworld}.}
\label{fig:diffusion_trans}
\vspace{-3mm}
\end{figure}

\subsection{Advanced Generation Models}
We also summarize existing generative models and how they are introduced in interactive world models.

\textbf{Variational Autoencoder (VAEs).} VAEs \cite{kingma2013auto} are designed by compressing data into a latent space with an encoder and reconstructing the original input by a decoder. VAEs leverage variational inference to approximate the posterior distribution over latent space. In the interactive world model domain, VAE is commonly used to compress spatial dimensions. For example, Genie \cite{bruce2024genie} designs a Latent Action Model (LAM) based on an encoder-decoder architecture to extract actions unsupervised from unlabeled videos.


\textbf{Generative Adversarial Networks (GANs).} GANs \cite{goodfellow2014generative} decompose the generation pipeline into two adversarial neural networks: a generator $G$ and a discriminator $D$. The generator creates synthetic data that mimics dataset distributions, while the discriminator distinguishes generated fake data samples. However, GANs are prone to mode collapse and limited generation diversity \cite{kim2020learning}.

\textbf{Diffusion Models (DMs).} Recently, diffusion models \cite{sohl2015deep,ho2020denoising} have gained tremendous advances, gradually removing noise from a Gaussian distribution to the data distribution by learning from its reverse diffusion process. Recent world models \cite{che2024gamegen,xiaoworldmem,dai2025fantasyworld,li2025magicworld,huang2026live,jiang2026olaf,nam2026worldcam} mostly rely on diffusion transformer (DiT) \cite{peebles2023scalable} with global receptive fields and enhanced scalability, as in Table \ref{tab:survey1}.

\textbf{Autoregressive Models (ARs).} Early video generation models establish no causal relationships among frames with bidirectional temporal cues in Fig. \ref{fig:diffusion_trans}. To model the causal world evolution, recent world models \cite{wu2024ivideogpt,guo2025mineworld,xiaoworldmem} pose a stronger constraint on causality in a frame-by-frame generation, where earlier generated frames can serve as conditions for later ones. Some recent methods, such as Vid2World \cite{huang2026vidworld}, specially design causality modules within the DiT network.

\setlength{\tabcolsep}{6mm}
\begin{table*}[t]
  \caption{\textbf{An overview of representative interactive world models in immersive game engines and open-world exploration fields.} Methods are compared in terms of input reference, world state, output, model architecture, user interfaces, application scenarios, and object editing ability. Here, \textit{DiT} indicates diffusion transformer, while \textit{diffusion model} means UNet-based or non-specified denoising networks. User actions are classified into: camera movement (\raisebox{-0.4ex}{\includegraphics[width=0.032\linewidth]{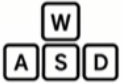}}), camera orientation (\raisebox{-0.4ex}{\includegraphics[width=0.032\linewidth]{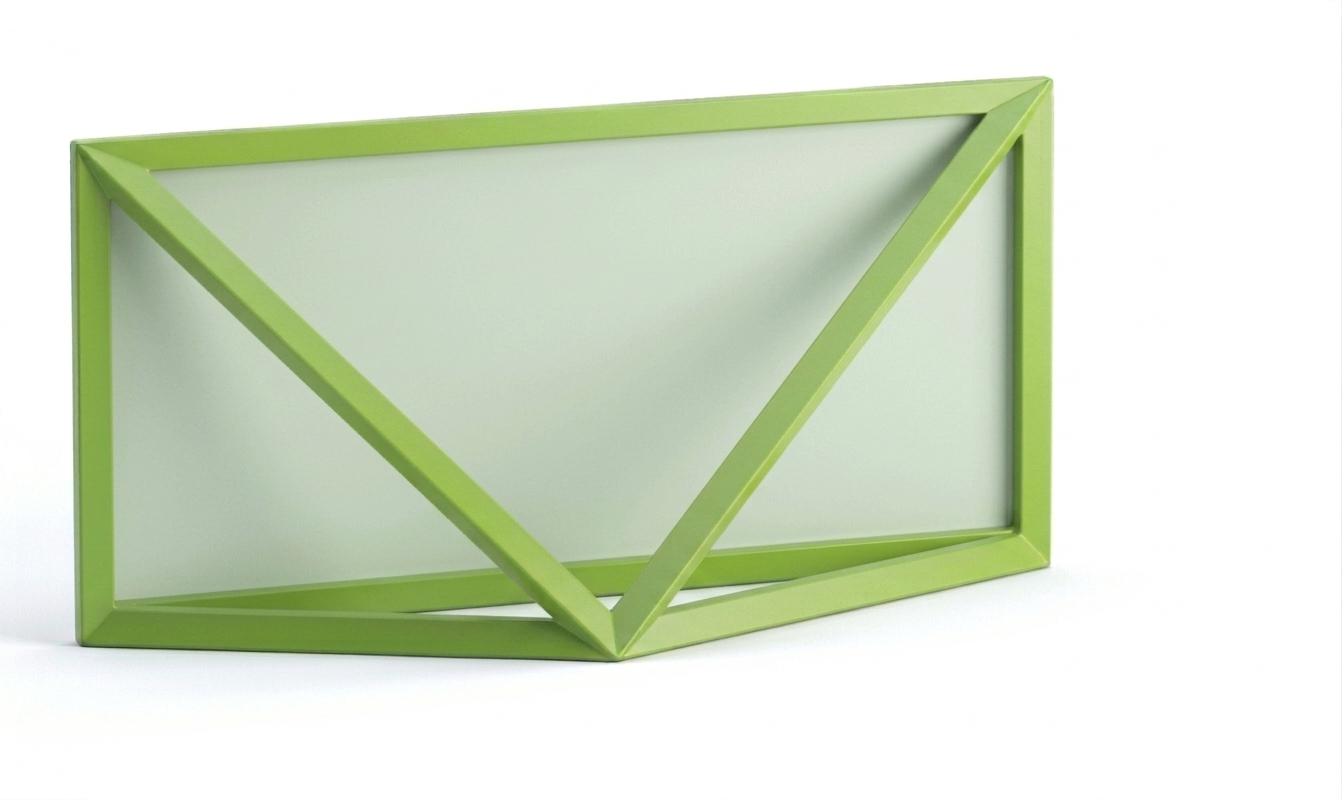}}), text (\raisebox{-0.4ex}{\includegraphics[width=0.014\linewidth]{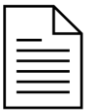}}), latent action (\raisebox{-0.4ex}{\includegraphics[width=0.022\linewidth]{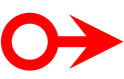}}), robot pose (\raisebox{-0.4ex}{\includegraphics[width=0.022\linewidth]{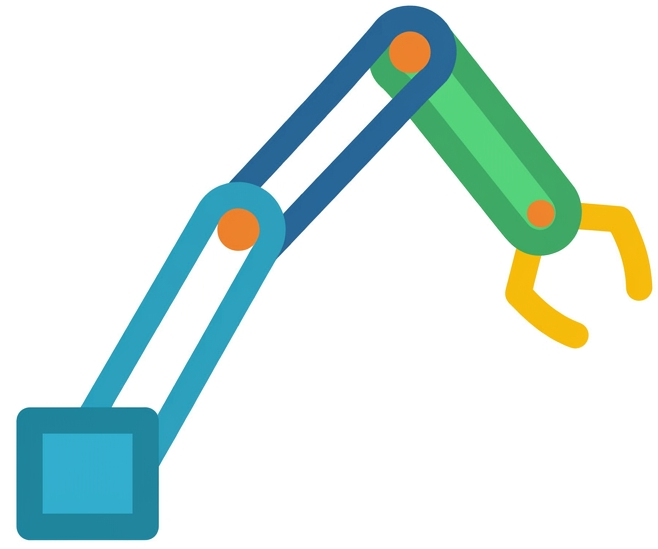}}), and object trajectory (\raisebox{-0.4ex}{\includegraphics[width=0.032\linewidth]{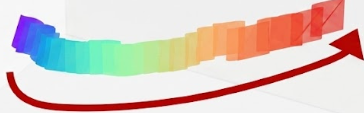}}). Applications are classified as game engines (\raisebox{-0.4ex}{\includegraphics[width=0.022\linewidth]{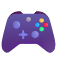}}), open-world exploration (\raisebox{-0.4ex}{\includegraphics[width=0.020\linewidth]{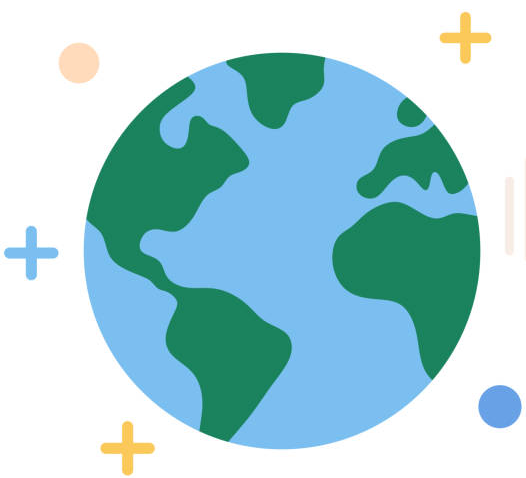}}), embodied AI (\raisebox{-0.4ex}{\includegraphics[width=0.020\linewidth]{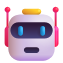}}), and autonomous driving (\raisebox{-0.4ex}{\includegraphics[width=0.025\linewidth]{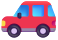}}).}
   \vspace{-0.8cm}
  \label{tab:survey1}
  \begin{center}
        \resizebox{\linewidth}{!}{
        \begin{tabular}{llccccccccc}
         \hline \toprule
           \textbf{Method} & \textbf{Venue} &\textbf{Input Reference} &\textbf{World State} & \textbf{Output Modality} &\textbf{Model Architecture}  & \textbf{User Interfaces} &\textbf{Application Scenarios} & \textbf{Object Editing}   \\    
          \midrule
          \rowcolor{myblue}GameGAN \cite{kim2020learning} & CVPR'20 &image &2D & video&GAN&\raisebox{-0.4ex}{\includegraphics[width=0.032\linewidth]{keyboard.png}} &\raisebox{-0.4ex}{\includegraphics[width=0.022\linewidth]{game.png}} & \xmark \\
          
          UniSim \cite{yang2024learning} & ICLR'24 & video & 2D & video & diffusion model&\raisebox{-0.4ex}{\includegraphics[width=0.014\linewidth]{text.png}} \raisebox{-0.4ex}{\includegraphics[width=0.032\linewidth]{pose.png}} \raisebox{-0.4ex}{\includegraphics[width=0.022\linewidth]{robopose.png}}& \raisebox{-0.4ex}{\includegraphics[width=0.020\linewidth]{world.png}} \raisebox{-0.4ex}{\includegraphics[width=0.020\linewidth]{robot.png}} \raisebox{-0.4ex}{\includegraphics[width=0.025\linewidth]{car.png}} & \cmark \\

          \rowcolor{myblue}Genie \cite{bruce2024genie}& ICML'24 & image, sketch & 2D & video & VAE &\raisebox{-0.4ex}{\includegraphics[width=0.022\linewidth]{latent.png}} & \raisebox{-0.4ex}{\includegraphics[width=0.022\linewidth]{game.png}} \raisebox{-0.4ex}{\includegraphics[width=0.020\linewidth]{world.png}} \raisebox{-0.4ex}{\includegraphics[width=0.020\linewidth]{robot.png}} & \cmark  \\

          Video2Game \cite{xia2024video2game}&CVPR'24 &video & 3D  & video & NeRF &\raisebox{-0.4ex}{\includegraphics[width=0.032\linewidth]{pose.png}} \raisebox{-0.4ex}{\includegraphics[width=0.032\linewidth]{keyboard.png}}& \raisebox{-0.4ex}{\includegraphics[width=0.022\linewidth]{game.png}} \raisebox{-0.4ex}{\includegraphics[width=0.020\linewidth]{world.png}} \raisebox{-0.4ex}{\includegraphics[width=0.020\linewidth]{robot.png}}& \cmark\\

         \rowcolor{myblue} WonderJourney \cite{yu2024wonderjourney}& CVPR'24 & image & 3D& point cloud video & diffusion model&\raisebox{-0.4ex}{\includegraphics[width=0.014\linewidth]{text.png}}  &\raisebox{-0.4ex}{\includegraphics[width=0.020\linewidth]{world.png}} &  \cmark\\

          DIAMOND \cite{alonso2024diffusion} & NeurIPS'24& video & 2D &video& diffusion model& \raisebox{-0.4ex}{\includegraphics[width=0.032\linewidth]{keyboard.png}} &\raisebox{-0.4ex}{\includegraphics[width=0.022\linewidth]{game.png}} &  \cmark \\

          \rowcolor{myblue}iVideoGPT \cite{wu2024ivideogpt} & NeurIPS'24 &image & 2D &video & autoregressive transformer  &\raisebox{-0.4ex}{\includegraphics[width=0.020\linewidth]{robopose.png}}  &\raisebox{-0.4ex}{\includegraphics[width=0.020\linewidth]{robot.png}} & \xmark\\


          Oasis \cite{oasis2024} &arxiv'24 &video &2D &video &DiT & \raisebox{-0.4ex}{\includegraphics[width=0.032\linewidth]{pose.png}} \raisebox{-0.4ex} {\includegraphics[width=0.032\linewidth]{keyboard.png}}  & \raisebox{-0.4ex}{\includegraphics[width=0.022\linewidth]{game.png}} & \xmark\\

          \rowcolor{myblue}GameGen-X \cite{che2024gamegen} & arxiv'24 &video&2D&video &DiT & \raisebox{-0.4ex}{\includegraphics[width=0.014\linewidth]{text.png}} \raisebox{-0.4ex}{\includegraphics[width=0.032\linewidth]{keyboard.png}}  & \raisebox{-0.4ex}{\includegraphics[width=0.022\linewidth]{game.png}} &\cmark \\

          Genie 2 \cite{parkerholder2024genie2} &2024 &image&2D&video&diffusion model&\raisebox{-0.4ex}{\includegraphics[width=0.014\linewidth]{text.png}}  \raisebox{-0.4ex}{\includegraphics[width=0.032\linewidth]{pose.png}} \raisebox{-0.4ex}{\includegraphics[width=0.032\linewidth]{keyboard.png}}&\raisebox{-0.4ex}{\includegraphics[width=0.022\linewidth]{game.png}} &\cmark \\

          \rowcolor{myblue}GameNGen \cite{valevski2025diffusion}&ICLR'25 & video &2D &video & diffusion model& \raisebox{-0.4ex}{\includegraphics[width=0.032\linewidth]{pose.png}} \raisebox{-0.4ex} {\includegraphics[width=0.032\linewidth]{keyboard.png}}  &\raisebox{-0.4ex}{\includegraphics[width=0.022\linewidth]{game.png}} & \xmark\\

          NWM \cite{bar2025navigation} &CVPR'25 &image &2D &video &DiT & \raisebox{-0.4ex}{\includegraphics[width=0.022\linewidth]{latent.png}}  & \raisebox{-0.4ex}{\includegraphics[width=0.020\linewidth]{robot.png}} &\xmark \\

          \rowcolor{myblue}WonderWorld \cite{yu2025wonderworld} &CVPR'25 &image &3D &3DGS &rendering & \raisebox{-0.4ex}{\includegraphics[width=0.014\linewidth]{text.png}} \raisebox{-0.4ex}{\includegraphics[width=0.032\linewidth]{pose.png}}&\raisebox{-0.4ex}{\includegraphics[width=0.020\linewidth]{world.png}} &\cmark \\

          AdaWorld \cite{gao2025adaworld} & ICML'25 & video &2D &video &diffusion model& \raisebox{-0.4ex}{\includegraphics[width=0.022\linewidth]{latent.png}}  & \raisebox{-0.4ex}{\includegraphics[width=0.022\linewidth]{game.png}} \raisebox{-0.4ex}{\includegraphics[width=0.020\linewidth]{world.png}} \raisebox{-0.4ex}{\includegraphics[width=0.020\linewidth]{robot.png}} & \cmark \\

          \rowcolor{myblue}WonderTurbo \cite{ni2025wonderturbo}& ICCV'25 & image &3D & 3DGS & rendering & \raisebox{-0.4ex}{\includegraphics[width=0.014\linewidth]{text.png}} \raisebox{-0.4ex}{\includegraphics[width=0.032\linewidth]{pose.png}}&\raisebox{-0.4ex}{\includegraphics[width=0.020\linewidth]{world.png}} &\cmark \\

          SSM-WM \cite{po2025long}& ICCV'25& video &2D &video &SSM+diffusion model& \raisebox{-0.4ex}{\includegraphics[width=0.032\linewidth]{keyboard.png}} &\raisebox{-0.4ex}{\includegraphics[width=0.022\linewidth]{game.png}} &\cmark\\

          \rowcolor{myblue}VMem \cite{li2025vmem}& ICCV'25& image &2D &video &diffusion model&\raisebox{-0.4ex}{\includegraphics[width=0.032\linewidth]{pose.png}} \raisebox{-0.4ex}{\includegraphics[width=0.032\linewidth]{keyboard.png}}  &\raisebox{-0.4ex}{\includegraphics[width=0.020\linewidth]{world.png}} & \xmark\\

          GameFactory \cite{Yu_2025_ICCV} &ICCV'25& video &2D &video &DiT & \raisebox{-0.4ex}{\includegraphics[width=0.032\linewidth]{pose.png}} \raisebox{-0.4ex}{\includegraphics[width=0.032\linewidth]{keyboard.png}} & \raisebox{-0.4ex}{\includegraphics[width=0.022\linewidth]{game.png}} & \xmark\\

          \rowcolor{myblue}AETHER \cite{zhu2025aether}&ICCV'25& image, video &4D &video &DiT & \raisebox{-0.4ex}{\includegraphics[width=0.032\linewidth]{pose.png}} \raisebox{-0.4ex}{\includegraphics[width=0.032\linewidth]{keyboard.png}}& \raisebox{-0.4ex}{\includegraphics[width=0.020\linewidth]{world.png}} \raisebox{-0.4ex}{\includegraphics[width=0.020\linewidth]{robot.png}} &\xmark\\

          WorldMem \cite{xiaoworldmem} &NeurIPS'25 & video &2D&video &DiT &\raisebox{-0.4ex}{\includegraphics[width=0.032\linewidth]{pose.png}} \raisebox{-0.4ex}{\includegraphics[width=0.032\linewidth]{keyboard.png}} &\raisebox{-0.4ex}{\includegraphics[width=0.022\linewidth]{game.png}} \raisebox{-0.4ex}{\includegraphics[width=0.020\linewidth]{world.png}} &\cmark\\

          \rowcolor{myblue}Spmem \cite{wu2025video} &NeurIPS'25 & video &3D&video&DiT& \raisebox{-0.4ex}{\includegraphics[width=0.014\linewidth]{text.png}} \raisebox{-0.4ex}{\includegraphics[width=0.032\linewidth]{pose.png}} &  \raisebox{-0.4ex}{\includegraphics[width=0.022\linewidth]{game.png}} \raisebox{-0.4ex}{\includegraphics[width=0.020\linewidth]{world.png}} &\cmark\\
          
          IaaW \cite{gui2025image}&NeurIPS'25 & image &2D&video &DiT&\raisebox{-0.4ex}{\includegraphics[width=0.032\linewidth]{pose.png}}  &\raisebox{-0.4ex}{\includegraphics[width=0.020\linewidth]{world.png}}&\cmark \\

          
          \rowcolor{myblue}The Matrix \cite{feng2025the} &NeurIPS'25 &video &2D&video&DiT &\raisebox{-0.4ex}{\includegraphics[width=0.014\linewidth]{text.png}} \raisebox{-0.4ex}{\includegraphics[width=0.032\linewidth]{keyboard.png}} & \raisebox{-0.4ex}{\includegraphics[width=0.022\linewidth]{game.png}} &\cmark\\



          MineWorld \cite{guo2025mineworld} &arxiv'25 &image&2D&video&autoregressive transformer & \raisebox{-0.4ex}{\includegraphics[width=0.032\linewidth]{pose.png}} \raisebox{-0.4ex}{\includegraphics[width=0.032\linewidth]{keyboard.png}}  & \raisebox{-0.4ex}{\includegraphics[width=0.022\linewidth]{game.png}} &\xmark  \\

          \rowcolor{myblue}DeepVerse \cite{chen2025deepverse}&arxiv'25&video, depth, raydrop &4D & point cloud video & DiT & \raisebox{-0.4ex}{\includegraphics[width=0.014\linewidth]{text.png}} &\raisebox{-0.4ex}{\includegraphics[width=0.022\linewidth]{game.png}} \raisebox{-0.4ex}{\includegraphics[width=0.020\linewidth]{world.png}}  &\cmark \\

          EmbodiedGen \cite{wang2025embodiedgen}&arxiv'25&image, text &3D &3DGS, mesh & diffusion model & \raisebox{-0.4ex}{\includegraphics[width=0.014\linewidth]{text.png}} & \raisebox{-0.4ex}{\includegraphics[width=0.020\linewidth]{world.png}} \raisebox{-0.4ex}{\includegraphics[width=0.020\linewidth]{robot.png}} &\cmark \\

          \rowcolor{myblue}HY-GameCraft \cite{li2025hunyuangamecrafthighdynamicinteractivegame}&arxiv'25&image, text&2D &video &DiT &\raisebox{-0.4ex}{\includegraphics[width=0.032\linewidth]{pose.png}} \raisebox{-0.4ex}{\includegraphics[width=0.032\linewidth]{keyboard.png}}  & \raisebox{-0.4ex}{\includegraphics[width=0.022\linewidth]{game.png}}&\cmark \\

          Matrix-Game \cite{zhang2025matrix} &arxiv'25 &image &2D &video &DiT & \raisebox{-0.4ex}{\includegraphics[width=0.032\linewidth]{pose.png}} \raisebox{-0.4ex}{\includegraphics[width=0.032\linewidth]{keyboard.png}} & \raisebox{-0.4ex}{\includegraphics[width=0.022\linewidth]{game.png}} &\xmark \\

           \rowcolor{myblue}HY-World 1.0 \cite{team2025hunyuanworld} &arxiv'25 &image, text &3D &mesh &DiT &\raisebox{-0.4ex}{\includegraphics[width=0.032\linewidth]{pose.png}} \raisebox{-0.4ex}{\includegraphics[width=0.032\linewidth]{keyboard.png}} &\raisebox{-0.4ex}{\includegraphics[width=0.022\linewidth]{game.png}} \raisebox{-0.4ex}{\includegraphics[width=0.020\linewidth]{world.png}}&\cmark\\

         Genie 3 \cite{genie3} &2025 &image&2D&video&autoregressive model&\raisebox{-0.4ex}{\includegraphics[width=0.014\linewidth]{text.png}} \raisebox{-0.4ex}{\includegraphics[width=0.032\linewidth]{pose.png}} \raisebox{-0.4ex}{\includegraphics[width=0.032\linewidth]{keyboard.png}} &\raisebox{-0.4ex}{\includegraphics[width=0.022\linewidth]{game.png}}\raisebox{-0.4ex}{\includegraphics[width=0.020\linewidth]{world.png}} \raisebox{-0.4ex}{\includegraphics[width=0.020\linewidth]{robot.png}} &\cmark \\

           \rowcolor{myblue}Matrix-3D \cite{yang2025matrix} &arxiv'25 &image, text &3D &mesh video &DiT &\raisebox{-0.4ex}{\includegraphics[width=0.032\linewidth]{pose.png}} \raisebox{-0.4ex}{\includegraphics[width=0.032\linewidth]{keyboard.png}} & \raisebox{-0.4ex}{\includegraphics[width=0.022\linewidth]{game.png}}&\xmark \\


          Matrix-game 2.0 \cite{he2025matrix}&arxiv'25 &image&2D &video&DiT & \raisebox{-0.4ex}{\includegraphics[width=0.032\linewidth]{pose.png}} \raisebox{-0.4ex}{\includegraphics[width=0.032\linewidth]{keyboard.png}}  & \raisebox{-0.4ex}{\includegraphics[width=0.022\linewidth]{game.png}} &\xmark\\

          \rowcolor{myblue}Dreamer-4 \cite{hafner2025training} &arxiv'25 &image&2D &video&autoregressive transformer & \raisebox{-0.4ex}{\includegraphics[width=0.032\linewidth]{pose.png}} \raisebox{-0.4ex}{\includegraphics[width=0.032\linewidth]{keyboard.png}}  & \raisebox{-0.4ex}{\includegraphics[width=0.022\linewidth]{game.png}} &\cmark\\

          Memory Forcing \cite{huang2025memory} &arxiv'25 &image&2D &video&DiT &\raisebox{-0.4ex}{\includegraphics[width=0.032\linewidth]{pose.png}} \raisebox{-0.4ex}{\includegraphics[width=0.032\linewidth]{keyboard.png}}  &\raisebox{-0.4ex}{\includegraphics[width=0.022\linewidth]{game.png}} &\xmark\\

         \rowcolor{myblue}PAN \cite{xiang2025pan}&arxiv'25 &image, video &2D &video &DiT  &\raisebox{-0.4ex}{\includegraphics[width=0.014\linewidth]{text.png}}  &\raisebox{-0.4ex}{\includegraphics[width=0.022\linewidth]{game.png}} \raisebox{-0.4ex}{\includegraphics[width=0.020\linewidth]{world.png}} \raisebox{-0.4ex}{\includegraphics[width=0.020\linewidth]{robot.png}}&\cmark \\

            MagicWorld \cite{li2025magicworld} &arxiv'25& image &2D &video &DiT &\raisebox{-0.4ex}{\includegraphics[width=0.032\linewidth]{keyboard.png}}  &\raisebox{-0.4ex}{\includegraphics[width=0.020\linewidth]{world.png}} &\xmark \\

           \rowcolor{myblue}RELIC \cite{hong2025relic} &arxiv'25 &image &2D &video &DiT &\raisebox{-0.4ex}{\includegraphics[width=0.014\linewidth]{text.png}} \raisebox{-0.4ex}{\includegraphics[width=0.032\linewidth]{pose.png}} \raisebox{-0.4ex}{\includegraphics[width=0.032\linewidth]{keyboard.png}} & \raisebox{-0.4ex}{\includegraphics[width=0.022\linewidth]{game.png}} \raisebox{-0.4ex}{\includegraphics[width=0.020\linewidth]{world.png}}  &\cmark \\


          WorldCanvas \cite{wang2025world}&arxiv'25 &image &2D &video&DiT & \raisebox{-0.4ex}{\includegraphics[width=0.014\linewidth]{text.png}}  \raisebox{-0.4ex}{\includegraphics[width=0.032\linewidth]{trajectory.png}} &\raisebox{-0.4ex}{\includegraphics[width=0.020\linewidth]{world.png}}&\cmark \\

          \rowcolor{myblue}TeleWorld \cite{chen2025teleworld}&arxiv'25 &video &4D &video&DiT & \raisebox{-0.4ex}{\includegraphics[width=0.032\linewidth]{keyboard.png}}&\raisebox{-0.4ex}{\includegraphics[width=0.020\linewidth]{world.png}} &\xmark \\ 

         Vid2World \cite{huang2026vidworld} &ICLR'26 &video &2D &video &DiT &\raisebox{-0.4ex}{\includegraphics[width=0.032\linewidth]{pose.png}} \raisebox{-0.4ex}{\includegraphics[width=0.032\linewidth]{keyboard.png}}  & \raisebox{-0.4ex}{\includegraphics[width=0.022\linewidth]{game.png}} \raisebox{-0.4ex}{\includegraphics[width=0.020\linewidth]{world.png}} \raisebox{-0.4ex}{\includegraphics[width=0.020\linewidth]{robot.png}} &\cmark \\

          \rowcolor{myblue}Astra \cite{zhu2026astra} & ICLR'26 &image &2D &video & DiT &\raisebox{-0.4ex}{\includegraphics[width=0.014\linewidth]{text.png}} \raisebox{-0.4ex}{\includegraphics[width=0.032\linewidth]{pose.png}} \raisebox{-0.4ex}{\includegraphics[width=0.032\linewidth]{keyboard.png}} &\raisebox{-0.4ex}{\includegraphics[width=0.020\linewidth]{world.png}} \raisebox{-0.4ex}{\includegraphics[width=0.020\linewidth]{robot.png}} \raisebox{-0.4ex}{\includegraphics[width=0.025\linewidth]{car.png}}&\cmark \\

         FantasyWorld \cite{dai2025fantasyworld}& ICLR'26 &image &3D &video & DiT & \raisebox{-0.4ex}{\includegraphics[width=0.014\linewidth]{text.png}} \raisebox{-0.4ex}{\includegraphics[width=0.032\linewidth]{pose.png}} \raisebox{-0.4ex}{\includegraphics[width=0.032\linewidth]{keyboard.png}} &\raisebox{-0.4ex}{\includegraphics[width=0.022\linewidth]{game.png}}\raisebox{-0.4ex}{\includegraphics[width=0.020\linewidth]{world.png}}&\cmark \\

          \rowcolor{myblue}MotionStream \cite{shin2026motionstream} & ICLR'26 &image &3D &video & DiT & \raisebox{-0.4ex}{\includegraphics[width=0.014\linewidth]{text.png}} \raisebox{-0.4ex}{\includegraphics[width=0.032\linewidth]{pose.png}} \raisebox{-0.4ex}{\includegraphics[width=0.032\linewidth]{keyboard.png}}  \raisebox{-0.4ex}{\includegraphics[width=0.032\linewidth]{trajectory.png}} & \raisebox{-0.4ex}{\includegraphics[width=0.020\linewidth]{world.png}} &\cmark \\

         NeoVerse \cite{yang2026neoverse} &CVPR'26 & image, video &4D &video &DiT &\raisebox{-0.4ex}{\includegraphics[width=0.014\linewidth]{text.png}} \raisebox{-0.4ex}{\includegraphics[width=0.032\linewidth]{pose.png}}  &\raisebox{-0.4ex}{\includegraphics[width=0.022\linewidth]{game.png}} \raisebox{-0.4ex}{\includegraphics[width=0.020\linewidth]{world.png}} \raisebox{-0.4ex}{\includegraphics[width=0.020\linewidth]{robot.png}} \raisebox{-0.4ex}{\includegraphics[width=0.025\linewidth]{car.png}} &\cmark \\

          \rowcolor{myblue}VerseCrafter \cite{zheng2026versecrafter} &CVPR'26 & image &4D &video &DiT &\raisebox{-0.4ex}{\includegraphics[width=0.032\linewidth]{pose.png}} \raisebox{-0.4ex}{\includegraphics[width=0.032\linewidth]{trajectory.png}} &\raisebox{-0.4ex}{\includegraphics[width=0.020\linewidth]{world.png}} &\cmark \\

          SonoWorld \cite{jin2026sonoworld}&CVPR'26 & image &3D &3D \& audio &rendering &\raisebox{-0.4ex}{\includegraphics[width=0.032\linewidth]{pose.png}} \raisebox{-0.4ex}{\includegraphics[width=0.032\linewidth]{keyboard.png}}  &\raisebox{-0.4ex}{\includegraphics[width=0.020\linewidth]{world.png}} &\xmark \\

          \rowcolor{myblue}PerpetualWonder \cite{zhan2026perpetualwonder} &CVPR'26 & image &4D &3DGS video &physical simulator &\raisebox{-0.4ex}{\includegraphics[width=0.022\linewidth]{latent.png}}  &\raisebox{-0.4ex}{\includegraphics[width=0.020\linewidth]{world.png}} &\cmark\\

          Wonderzoom \cite{cao2025wonderzoom}&CVPR'26&image & 3D &3DGS &rendering & \raisebox{-0.4ex}{\includegraphics[width=0.014\linewidth]{text.png}} \raisebox{-0.4ex}{\includegraphics[width=0.032\linewidth]{pose.png}}  &\raisebox{-0.4ex}{\includegraphics[width=0.020\linewidth]{world.png}} &\cmark \\

             \rowcolor{myblue}WorldGen \cite{wang2025worldgen}&CVPR'26&image, text & 3D &mesh &DiT & \raisebox{-0.4ex}{\includegraphics[width=0.014\linewidth]{text.png}}  &\raisebox{-0.4ex}{\includegraphics[width=0.022\linewidth]{game.png}} &\cmark \\

         Yume \cite{mao2025yume1,mao2025yume1.5}&CVPR'26& image, video, text & 2D &video &DiT &\raisebox{-0.4ex}{\includegraphics[width=0.032\linewidth]{pose.png}} \raisebox{-0.4ex}{\includegraphics[width=0.032\linewidth]{keyboard.png}} &\raisebox{-0.4ex}{\includegraphics[width=0.020\linewidth]{world.png}}&\cmark \\

          \rowcolor{myblue}OmniRoam \cite{liu2026omniroam}&SIGGRAPH'26& image, video &3D &video &rectified flow &\raisebox{-0.4ex}{\includegraphics[width=0.032\linewidth]{pose.png}} \raisebox{-0.4ex}{\includegraphics[width=0.032\linewidth]{keyboard.png}}&\raisebox{-0.4ex}{\includegraphics[width=0.020\linewidth]{world.png}} & \xmark\\

           HY-WorldPlay \cite{sun2025worldplay} &ICML'26 &image, text&2D &video&DiT & \raisebox{-0.4ex}{\includegraphics[width=0.032\linewidth]{pose.png}} \raisebox{-0.4ex}{\includegraphics[width=0.032\linewidth]{keyboard.png}} & \raisebox{-0.4ex}{\includegraphics[width=0.022\linewidth]{game.png}} \raisebox{-0.4ex}{\includegraphics[width=0.020\linewidth]{world.png}} &\xmark\\

           \rowcolor{myblue}LIVE \cite{huang2026live} &ICML'26 & video &2D &video &DiT & \raisebox{-0.4ex}{\includegraphics[width=0.032\linewidth]{pose.png}} \raisebox{-0.4ex}{\includegraphics[width=0.032\linewidth]{keyboard.png}} & \raisebox{-0.4ex}{\includegraphics[width=0.022\linewidth]{game.png}} \raisebox{-0.4ex}{\includegraphics[width=0.020\linewidth]{world.png}} &\xmark\\

            Infinite-World \cite{wu2026infinite}&ICML'26 & video &2D &video &DiT & \raisebox{-0.4ex}{\includegraphics[width=0.032\linewidth]{keyboard.png}}  &\raisebox{-0.4ex}{\includegraphics[width=0.020\linewidth]{world.png}} &\xmark\\

            \rowcolor{myblue}WorldCompass \cite{wang2026worldcompass} &ICML'26 & video &2D &video &DiT & \raisebox{-0.4ex}{\includegraphics[width=0.032\linewidth]{keyboard.png}} &\raisebox{-0.4ex}{\includegraphics[width=0.020\linewidth]{world.png}} &\xmark\\

            Olaf-World \cite{jiang2026olaf}&ICML'26 & video &2D &video &DiT & \raisebox{-0.4ex}{\includegraphics[width=0.022\linewidth]{latent.png}}  &\raisebox{-0.4ex}{\includegraphics[width=0.020\linewidth]{world.png}} &\xmark\\

            \rowcolor{myblue}SphericalDreamer \cite{schnepf2026sphericaldreamer}&ICML'26 & text &3D &point cloud &rendering & \raisebox{-0.4ex}{\includegraphics[width=0.032\linewidth]{pose.png}} \raisebox{-0.4ex}{\includegraphics[width=0.032\linewidth]{keyboard.png}}  &\raisebox{-0.4ex}{\includegraphics[width=0.020\linewidth]{world.png}} &\xmark\\




          EgoWM \cite{bagchi2026walk} &arxiv'26 &image &2D &video &DiT & \raisebox{-0.4ex}{\includegraphics[width=0.032\linewidth]{pose.png}} \raisebox{-0.4ex}{\includegraphics[width=0.032\linewidth]{keyboard.png}} \raisebox{-0.4ex}{\includegraphics[width=0.022\linewidth]{robopose.png}}  & \raisebox{-0.4ex}{\includegraphics[width=0.020\linewidth]{world.png}} \raisebox{-0.4ex}{\includegraphics[width=0.020\linewidth]{robot.png}} &\cmark\\

           \rowcolor{myblue}StableWorld \cite{yang2026stableworld} &arxiv'26 &image &2D &video &DiT &\raisebox{-0.4ex}{\includegraphics[width=0.032\linewidth]{keyboard.png}}  & \raisebox{-0.4ex}{\includegraphics[width=0.022\linewidth]{game.png}}&\xmark \\

          LingBot-World \cite{team2026advancing} &arxiv'26 &image, video &2D &video &DiT &\raisebox{-0.4ex}{\includegraphics[width=0.014\linewidth]{text.png}}  \raisebox{-0.4ex}{\includegraphics[width=0.032\linewidth]{keyboard.png}}  & \raisebox{-0.4ex}{\includegraphics[width=0.022\linewidth]{game.png}} \raisebox{-0.4ex}{\includegraphics[width=0.020\linewidth]{world.png}}&\cmark \\



           \rowcolor{myblue}Solaris \cite{savva2026solaris}&arxiv'26 &image&2D &video&DiT & \raisebox{-0.4ex}{\includegraphics[width=0.032\linewidth]{pose.png}} \raisebox{-0.4ex}{\includegraphics[width=0.032\linewidth]{keyboard.png}}  & \raisebox{-0.4ex}{\includegraphics[width=0.022\linewidth]{game.png}} &\cmark\\

          PERSIST \cite{garcin2026beyond}&arxiv'26 &image&3D &video&DiT & \raisebox{-0.4ex}{\includegraphics[width=0.032\linewidth]{pose.png}} \raisebox{-0.4ex}{\includegraphics[width=0.032\linewidth]{keyboard.png}}  & \raisebox{-0.4ex}{\includegraphics[width=0.022\linewidth]{game.png}} &\xmark\\

           \rowcolor{myblue}RealWonder \cite{liu2026realwonder} &arxiv'26 &image&3D &video&DiT &\raisebox{-0.4ex}{\includegraphics[width=0.022\linewidth]{latent.png}}  &\raisebox{-0.4ex}{\includegraphics[width=0.020\linewidth]{world.png}} \raisebox{-0.4ex}{\includegraphics[width=0.020\linewidth]{robot.png}} &\cmark\\



           LiveWorld \cite{duan2026liveworld} &arxiv'26 &video&4D &video&DiT &\raisebox{-0.4ex}{\includegraphics[width=0.032\linewidth]{pose.png}} \raisebox{-0.4ex}{\includegraphics[width=0.032\linewidth]{keyboard.png}} &\raisebox{-0.4ex}{\includegraphics[width=0.020\linewidth]{world.png}} &\xmark\\

             \rowcolor{myblue}SWM \cite{seo2026grounding} &arxiv'26 &video&2D &video&DiT &  \raisebox{-0.4ex}{\includegraphics[width=0.014\linewidth]{text.png}} \raisebox{-0.4ex}{\includegraphics[width=0.032\linewidth]{pose.png}} \raisebox{-0.4ex}{\includegraphics[width=0.032\linewidth]{keyboard.png}}&\raisebox{-0.4ex}{\includegraphics[width=0.020\linewidth]{world.png}} &\cmark\\

         WorldCam \cite{nam2026worldcam}&arxiv'26 &image, video&3D &video&DiT & \raisebox{-0.4ex}{\includegraphics[width=0.014\linewidth]{text.png}} \raisebox{-0.4ex}{\includegraphics[width=0.032\linewidth]{pose.png}} \raisebox{-0.4ex}{\includegraphics[width=0.032\linewidth]{keyboard.png}}  &\raisebox{-0.4ex}{\includegraphics[width=0.022\linewidth]{game.png}}&\cmark\\


           \rowcolor{myblue}MosaicMem \cite{yu2026mosaicmem}& arxiv'26 &image&2D & video&DiT &\raisebox{-0.4ex}{\includegraphics[width=0.014\linewidth]{text.png}} \raisebox{-0.4ex}{\includegraphics[width=0.032\linewidth]{pose.png}} \raisebox{-0.4ex}{\includegraphics[width=0.032\linewidth]{keyboard.png}}   &\raisebox{-0.4ex}{\includegraphics[width=0.020\linewidth]{world.png}} &\cmark\\

          ActionParty \cite{pondaven2026actionparty} &arxiv'26 &video&2D &video&DiT & \raisebox{-0.4ex}{\includegraphics[width=0.032\linewidth]{pose.png}} \raisebox{-0.4ex}{\includegraphics[width=0.032\linewidth]{keyboard.png}} & \raisebox{-0.4ex}{\includegraphics[width=0.022\linewidth]{game.png}}&\cmark\\

           \rowcolor{myblue}INSPATIO-WORLD \cite{team2026inspatio} &arxiv'26 &video&4D &video&DiT & \raisebox{-0.4ex}{\includegraphics[width=0.032\linewidth]{pose.png}} \raisebox{-0.4ex}{\includegraphics[width=0.032\linewidth]{keyboard.png}}  &\raisebox{-0.4ex}{\includegraphics[width=0.020\linewidth]{world.png}}  &\xmark\\

          Matrix-Game 3.0 \cite{matrixgame3} &arxiv'26 &video&2D &video&DiT & \raisebox{-0.4ex}{\includegraphics[width=0.032\linewidth]{pose.png}} \raisebox{-0.4ex}{\includegraphics[width=0.032\linewidth]{keyboard.png}}  &\raisebox{-0.4ex}{\includegraphics[width=0.022\linewidth]{game.png}}&\xmark\\

           \rowcolor{myblue}Lyra 2.0 \cite{shen2026lyra2} &arxiv'26 &image &3D &3DGS, point cloud &DiT & \raisebox{-0.4ex}{\includegraphics[width=0.032\linewidth]{pose.png}} \raisebox{-0.4ex}{\includegraphics[width=0.032\linewidth]{keyboard.png}} &\raisebox{-0.4ex}{\includegraphics[width=0.022\linewidth]{game.png}} \raisebox{-0.4ex}{\includegraphics[width=0.020\linewidth]{world.png}} \raisebox{-0.4ex}{\includegraphics[width=0.020\linewidth]{robot.png}}&\cmark\\

         HY-World 2.0 \cite{hy2026hy} &arxiv'26 &image, video, text &3D &3DGS, mesh &DiT & \raisebox{-0.4ex}{\includegraphics[width=0.014\linewidth]{text.png}}
          \raisebox{-0.4ex}{\includegraphics[width=0.032\linewidth]{pose.png}} \raisebox{-0.4ex}{\includegraphics[width=0.032\linewidth]{keyboard.png}} &\raisebox{-0.4ex}{\includegraphics[width=0.022\linewidth]{game.png}} \raisebox{-0.4ex}{\includegraphics[width=0.020\linewidth]{world.png}} \raisebox{-0.4ex}{\includegraphics[width=0.020\linewidth]{robot.png}}&\cmark\\

           \rowcolor{myblue}Happy Oyster \cite{happy} &2026 &image, video, text, audio &--- &--- &--- &\raisebox{-0.4ex}{\includegraphics[width=0.014\linewidth]{text.png}}  \raisebox{-0.4ex}{\includegraphics[width=0.032\linewidth]{pose.png}} \raisebox{-0.4ex}{\includegraphics[width=0.032\linewidth]{keyboard.png}} &\raisebox{-0.4ex}{\includegraphics[width=0.022\linewidth]{game.png}} \raisebox{-0.4ex}{\includegraphics[width=0.020\linewidth]{world.png}}&\cmark\\

          MultiWorld \cite{wu2026multiworld}&arxiv'26 &image &2D &video &DiT & \raisebox{-0.4ex}{\includegraphics[width=0.014\linewidth]{text.png}}
          \raisebox{-0.4ex}{\includegraphics[width=0.032\linewidth]{pose.png}} \raisebox{-0.4ex}{\includegraphics[width=0.032\linewidth]{keyboard.png}} &\raisebox{-0.4ex}{\includegraphics[width=0.022\linewidth]{game.png}} \raisebox{-0.4ex}{\includegraphics[width=0.020\linewidth]{world.png}}&\cmark\\

           \rowcolor{myblue}SANA-WM \cite{zhu2026sana} &arxiv'26 &multi-view image &2D &video &DiT & \raisebox{-0.4ex}{\includegraphics[width=0.014\linewidth]{text.png}}
         \raisebox{-0.4ex}{\includegraphics[width=0.032\linewidth]{keyboard.png}} &\raisebox{-0.4ex}{\includegraphics[width=0.022\linewidth]{game.png}} \raisebox{-0.4ex}{\includegraphics[width=0.020\linewidth]{robot.png}}&\cmark\\
          

          \bottomrule \hline
        \end{tabular}
        }
  
  \end{center}
  \vspace{-0.8cm}
\end{table*}

\section{Recent Research Trends}
\label{sec:trend}
In this section, we review recent research trends in the interactive world modeling domain, providing a general understanding of how advanced this field is and which specific fields researchers currently focus on.

\subsection{Applicability: From Specialist to Generalist} 
As in Table \ref{tab:survey1}, early methods are typically developed for a single application domain, such as game engine \cite{oasis2024,che2024gamegen,team2025hunyuanworld,zhang2025matrix,he2025matrix,mao2025yume1,guo2025mineworld,xia2024video2game,bruce2024genie,parkerholder2024genie2,genie3}, open-world exploration \cite{yu2025wonderworld,yu2024wonderjourney,cao2025wonderzoom,ni2025wonderturbo,zheng2026versecrafter}, etc. To develop world models that generalize across diverse domains, recent methods leverage large-scale pre-trained video backbones such as Wan 2.2 \cite{wan2025wan}, leveraging knowledge acquired from large-scale Internet video corpora. UniSim \cite{yang2024learning} enhances the generalization ability by mixing different datasets for training, including simulated environments, real-world robot data, human activity data, and panorama scans. AdaWorld \cite{gao2025adaworld} extracts latent actions from unlabeled videos in a self-supervised manner and integrates these learned actions into the pre-training procedure. This strategy disentangles the most crucial actions from in-the-wild videos, enabling adaptive action transfer across contexts. Astra \cite{zhu2026astra} proposes a mixture of action experts to dynamically route heterogeneous modalities, with improved versatility in open-world exploration, robotics, and autonomous driving. GAIA-1~\cite{hu2023gaia1} and DriveDreamer~\cite{wang2024drivedreamer} unify different tasks like scene perception and planning within a shared generative framework. In addition, recent industry-level products, such as Genie \cite{bruce2024genie,parkerholder2024genie2,genie3}, HY-World \cite{team2025hunyuanworld,hy2026hy}, Happy Oyster \cite{happy}, Marble \cite{marble}, LingBot-World \cite{team2026advancing}, etc, mostly support generation in versatile scenarios.


\begin{figure}[t]
\centering
\includegraphics[width=1.0\linewidth]{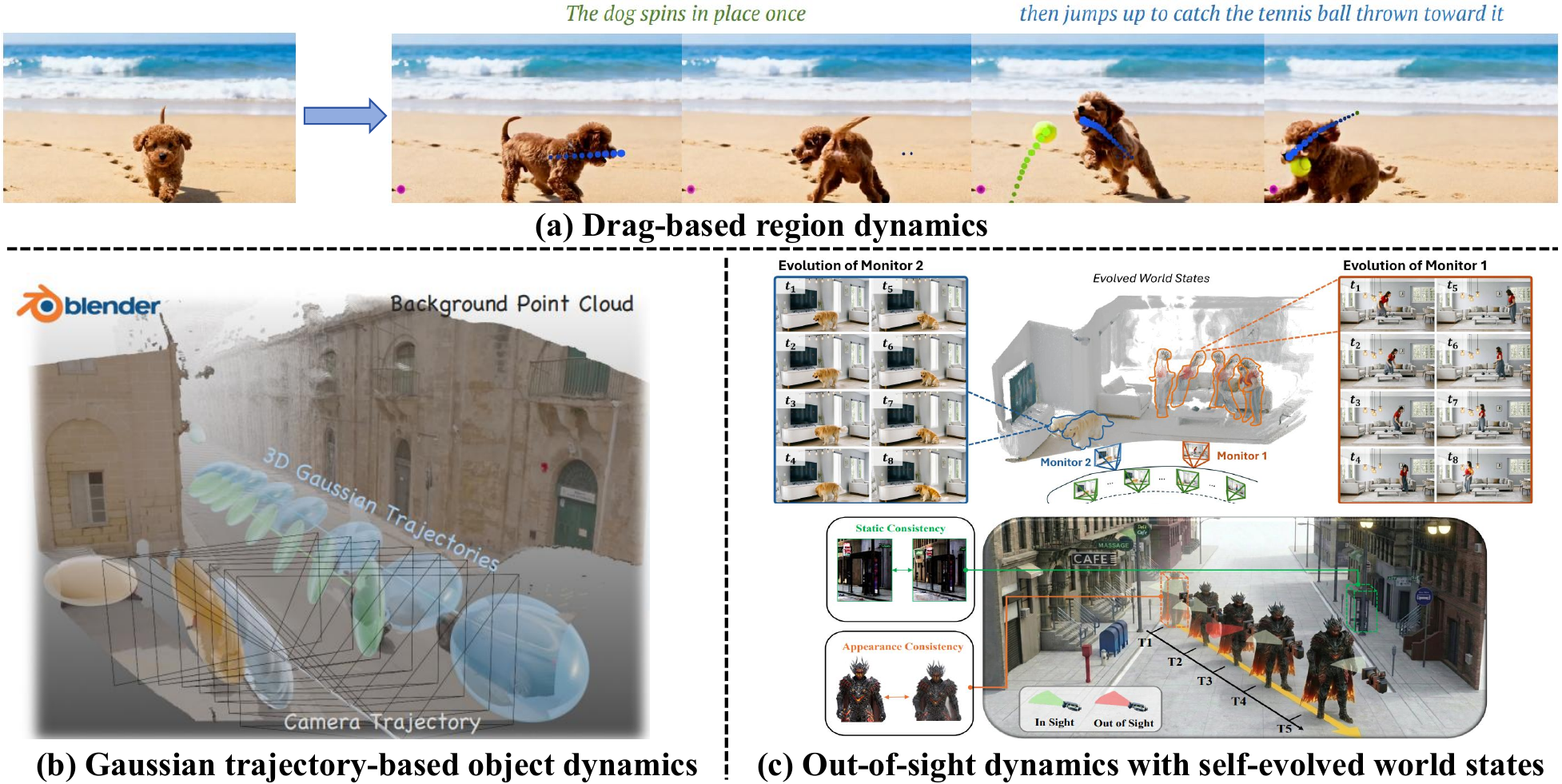}
\vspace{-6mm}
\caption{\textbf{Comparison of different levels of dynamics.} WorldCanvas \cite{wang2025world} introduces region-level dynamics instructed by text and drag gesture. VerseCrafter \cite{zheng2026versecrafter} models per-object dynamics by Gaussian trajectories in 4D geometric control. LiveWorld \cite{duan2026liveworld} even extends from in-sight-only dynamics to out-of-sight ones, enabling synchronized world evolution.}
\label{fig:dynamics}
\vspace{-3mm}
\end{figure} 

\subsection{World State: From Static, Single-Agent to Dynamic, Self-Evolving, Multi-Agent World} 
\textbf{From Static-only to Dynamic, Self-evolving Scenarios.} 
Early methods \cite{yu2025wonderworld,yu2024wonderjourney,wang2025worldgen} generate explorable but temporally static environments, where dynamic object interactions are absent. For example, in WonderWorld \cite{yu2025wonderworld}, users can impose camera motion commands to traverse the contents of the generated scene. However, the generated environments remain temporally frozen after each interaction. Some recent methods \cite{wang2026worldcompass,sun2025worldplay,yang2026neoverse,zheng2026versecrafter,team2026inspatio,zhan2026perpetualwonder,garcin2026beyond} start to incorporate dynamic entities as in Fig. \ref{fig:dynamics}. WorldCanvas \cite{wang2025world} and MotionStream \cite{shin2026motionstream} enable both object-level motion grounding and trajectory customization through a 'drag-and-drop' interface. NeoVerse \cite{yang2026neoverse} and VerseCrafter \cite{zheng2026versecrafter} propose novel 4D modeling pipelines by introducing per-object trajectories. Nevertheless, the generated world only evolves in the observer's field of view (FOV) in these models, commonly neglecting the fact that unobserved regions continue to evolve simultaneously. As a consequence, revisited regions may exhibit inconsistent world states. To address the 'out-of-sight dynamic' issue, LiveWorld \cite{duan2026liveworld} establishes a global self-evolving state to simulate the temporal progression. HM-World \cite{chen2026out} further improves the fine-grained motion modeling of dynamic objects by developing a hybrid memory and vigilant trackers. 

\textbf{From Single-agent to Multi-agent Interaction.} 
Another advancement lies in the higher complexity of the interactive characters. Most methods only consider single-agent setting, where the main observer wanders while outpainting extended world \cite{ni2025wonderfree}. However, this formulation is too simplified, which cannot reflect the environment complexity in real worlds. To model multi-agent involvement, Versecrafter \cite{zheng2026versecrafter} disentangles static backgrounds and dynamic foreground objects, which are rendered by point cloud and per-object 4D Gaussian Splatting, respectively. Solaris \cite{savva2026solaris} proposes a multi-player world model, in which a shared self attention mechanism is deployed to capture cross-agent feature exchange. Similarly, MultiGen \cite{po2026multigen},
AgentParty \cite{pondaven2026actionparty}, and MultiWorld \cite{wu2026multiworld} establish immersive game interactions with multiple agents. Combo \cite{zhang2024combo} designs a compositional world model for embodied multi-agent cooperation given by per-agent partial egocentric observations. ShareVerse \cite{zhu2026shareverse} further curates a novel multi-agent dataset based on the CARLA simulator, and also propose cross-view spatial concatenation and cross-attention agent interaction for video pre-training.  

\subsection{Interactive Modality: From Single-Sensory to Multi-Sensory Interfaces}
Humans understand and predict the future far beyond visual-only perception, where counterfactual reasoning ability arises from mixed sensory inputs, including audio, tactile, force, gravity, etc. Therefore, it is crucial to integrate heterogeneous sensory signals to achieve general spatial intelligence. Recently, SonoWorld \cite{jin2026sonoworld} extends the classic image-to-3D world exploration by integrating additional audio output. Another line of research \cite{liu2026realwonder, zhan2026perpetualwonder} investigates merging physic-informed actions, such as force, gravity field, etc, as conditions to guide physics-aware future predictions. Compared with open-world settings, autonomous driving community highlights more structured interaction interfaces, such as ego velocity, signals from surrounding agents, road semantics, High-Definition Maps (HD Map), and synchronized multi-camera observations~\cite{wang2024drivewm,russell2025gaia2}.

\section{User Action Controllability}
\label{sec:control}
The primary distinction between interactive world models and conventional video generation models lies in action-conditioned controllability. As in Fig. \ref{fig:compare_control}, early video generation models, such as ControlNet \cite{zhang2023adding}, adopt a one-shot text or visual instruction, lacking multi-round interaction from users. In contrast, interactive world models support iterative user interactions, enabling progressively refined and fine-grained control over world evolution \cite{sun2025worldplay,ni2025wonderturbo}. Furthermore, researchers have further explored diverse action-conditioning mechanisms tailored to different input modalities..



\subsection{One-Shot Control in Video Generation}
Recent advances in diffusion and autoregressive generative models have significantly improved text-to-video \cite{singer2022make,henschel2025streamingt2v} and image-to-video \cite{yu2024viewcrafter,cao2025uni3c} generation capabilities. However, text and image prompt-based conditioning offers only coarse-grained control over future generation outcomes. As in Fig. \ref{fig:compare_control} (up), text instructions are used as conditions for all generated frames in a one-shot manner. This weakens temporal causality and also lacks step-by-step supervision. 

Another line of work \cite{yu2024viewcrafter, cao2025uni3c,wang2024motionctrl} formulates conditional video generation with a camera-controlled 3D reconstruction pipeline, where an initial image is given and then a predefined camera trajectory is used to synthesize novel viewpoints. ViewCrafter \cite{yu2024viewcrafter} enables zero-shot novel-view synthesis by leveraging pretrained diffusion priors. MotionCtrl \cite{wang2024motionctrl} further independently decomposes camera and object motions. Uni3C \cite{cao2025uni3c} proposes a unified plug-and-play controller for both camera and human control and also jointly aligned 3D world guidance to ensure scene-character consistency. Subsequent approaches \cite{watson2024controlling,ren2025gen3c,chen2026efficient} improve view synthesis quality by introducing additional 3D consistency or temporal supervision, or increase generation speed. Although these methods can let users pre-define a camera trajectory to sample imagined views, interaction terminates once the conditioning signal has been specified. Furthermore, these methods are largely restricted by a small range of extrapolations, with limited ability to create entirely novel scenarios due to their reliance on view reconstruction rather than world-state evolution.

\subsection{Multi-Shot Fine-Grained Control in World Models}

Unlike general video generation models, recent interactive world models \cite{xiaoworldmem,li2025magicworld,jiang2026olaf,team2025hunyuanworld} adopt a per-frame action-conditioned paradigm, which enables users to provide fine-grained guidance to control immediate generation results in real time as in Fig. \ref{fig:compare_control} (bottom). 


Early world models like Dreamer \cite{hafner2019dream} and MuZero \cite{schrittwieser2020mastering} rely on recurrent models like RNN to provide step-level interactivity. However, because of the nature of RNN, their scalability remains limited. Recent world models leverage the autoregressive model to encode action inputs into causally ordered token sequences, enabling both high scalability while preserving \textit{step-level} interactivity in Fig. \ref{fig:diffusion_trans}. iVideoGPT \cite{wu2024ivideogpt} first proposes an autoregressive transformer for interaction video prediction. UniSim \cite{yang2024learning} and its followers commonly adopt an autoregressive diffusion model, predicting future tokens conditioned on historical observations and current \textit{frame-level} actions. Some recent methods \cite{wang2025world,yang2026neoverse,zheng2026versecrafter} even increase the controllability supporting \textit{region-level} or even \textit{object-level} interactions. Among these, WorldCanvas \cite{wang2025world} detects foreground objects as movable agents, allowing users to intuitively edit the selected agents with a `drag-and-drop' manner to animate each instance. NeoVerse \cite{yang2026neoverse} combines degraded rendering attributes from a feed-forward network VGGT \cite{wang2025vggt} into a control branch, thereby enabling dynamic 4D world modeling. VerseCrafter \cite{zheng2026versecrafter} further designs a 4D geometric control interface, decomposing the scene with a static point cloud and multi-object dynamics with per-object 3D Gaussian trajectories. This disentanglement offers the network greater flexibility to model more fine-grained dynamics.

\begin{figure*}[t]
\centering
\includegraphics[width=1.0\linewidth]{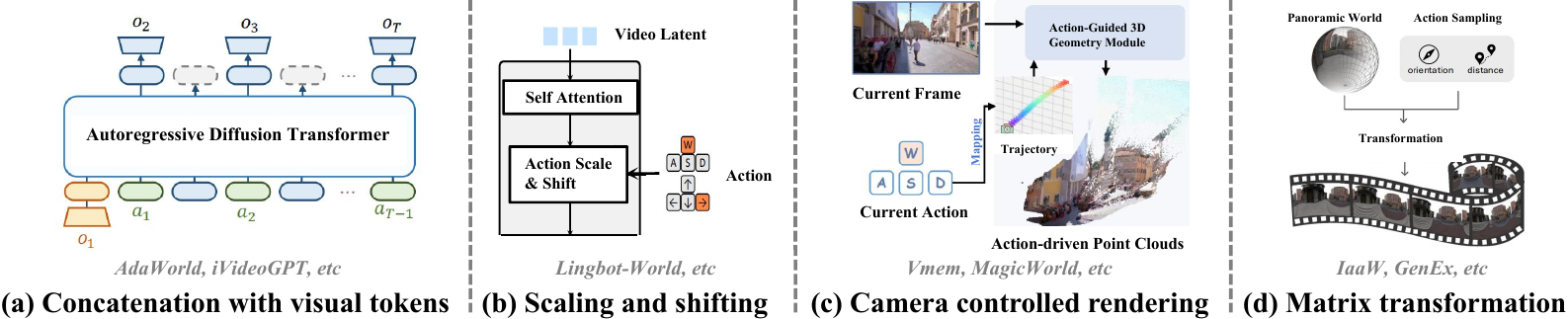}
\vspace{-6mm}
\caption{\textbf{Comparison of various injection manners for camera actions.} We classify main injection manners into four categories: (a) Concatenation with visual tokens \cite{wu2024ivideogpt}; (b) Scaling and Shifting \cite{team2026advancing}; (c) Camera controlled rendering or simulation \cite{li2025magicworld}; and (d) Matrix transformation \cite{lu2024genex}.}
\label{fig:action_injection}
\vspace{-3mm}
\end{figure*}

\subsection{Diverse Action Injection Manners}
Corresponding to different action inputs, recent works design unique injection interfaces to better guide future imaginations and fully capture intrinsic signal characteristics. Among these, two categories of actions are commonly used in existing interactive world models: camera movement or orientation \cite{yu2025wonderworld,ni2025wonderturbo} and text \cite{chen2025deepverse,xiang2025pan,wang2025worldgen} as in Table \ref{tab:survey1}. Typically, the camera signal controls moving motions and character trajectories, exerting minimal impact on overall scenarios \cite{che2024gamegen}, and there are basically high consistent rollouts across frames. In contrast, textual instructions govern higher-level semantic attributes of the environment, such as lighting conditions, styles, and can add/remove novel agents within the environment, significantly affecting future predictions and also offering higher-level customization from users. Therefore, we separately describe the action injection manners for each modality to provide a better understanding of relevant technique details.

\subsubsection{Injection of Camera Pose or Trajectory}
Camera orientations and movements are particularly crucial in creating interactive open-world exploration \cite{ni2025wonderturbo,yu2025wonderworld} and game engine \cite{hong2025relic,he2025matrix,mao2025yume1}, allowing users to freely traverse (controlled by the camera trajectory or keyboard) and look around (controlled by the camera pose, keyboard, or mouse) within the generated worlds or games. As in Fig. \ref{fig:action_injection}, there are typically four main methods to inject camera actions:

\textbf{(1) Concatenation with Visual Tokens as Conditions.} A natural injection manner is to lift the dimension of the camera-related inputs and then concatenate them with the denoised video tokens. As a pioneering work, Genie \cite{bruce2024genie} designs a Dynamics Model to recover masked future videos from previous video frames and user actions. During inference, future video clips are autoregressively predicted and combined with action tokens as the input of the Dynamics Model. iVideoGPT \cite{wu2024ivideogpt} proposes a special slot token to indicate frame-level boundaries and facilitate the concatenation of low-dimension action tokens. AdaWorld \cite{gao2025adaworld} also introduces a latent action-aware pre-training strategy from unlabeled videos, where latent actions can be transferred and reused in different contexts. Subsequent approaches \cite{huang2025memory,wu2025video,guo2025mineworld,li2025magicworld,huang2026vidworld,yang2026neoverse,wu2026infinite,wang2026worldcompass} usually adopt a similar concatenation manner. GameFactory \cite{Yu_2025_ICCV} and the Matrix-Game series \cite{zhang2025matrix,he2025matrix,matrixgame3} further distinguish keyboard and mouse input, modeling mouse movements as continuous scalar values with changing pitch angles while leveraging discrete embeddings to indicate keyboard inputs, such as 'up and down', 'jump', or 'attack'. Then, continuous mouse movements are concatenated with visual latent tokens for subsequent temporal attention. In contrast, discrete keyboard inputs are integrated by cross attention.

\textbf{(2) Scaling and Shifting Visual Tokens.} As in Fig. \ref{fig:action_injection}, some methods \cite{che2024gamegen,xiaoworldmem,sun2025worldplay,dai2025fantasyworld,team2026advancing,po2025long} modulate latent visual features by scaling and shifting processes. For example, GameGen-X \cite{che2024gamegen} designs multi-modal experts to ensure that each action signal is well utilized, where scale and shift parameters are generated from a neural network prompted by keyboard inputs.

\textbf{(3) Camera-Controlled Rendering or Simulation.} There is also another line of research \cite{li2025vmem,huang2025memory,duan2026liveworld,seo2026grounding} which reframes future video generation task as camera-controlled reconstruction. Among these, Vmem \cite{li2025vmem} proposes a Surfel-index view memory to store past views and retrieve the most similar one through rendering given a novel target camera pose in Fig. \ref{fig:action_injection}. LiveWorld \cite{duan2026liveworld} reconstructs the static background with a SLAM system and projects evolved foreground entities and background environment by rendering novel views to generate future states. SWM \cite{seo2026grounding} grounds the simulated world in a real city, where the pre-stored spatially nearest reference images are retrieved and re-projected into the target viewpoint via a depth-guided splatting algorithm. Some methods even resort to webGL-based game engines \cite{xia2024video2game} or existing physical simulators \cite{liu2026realwonder} to generate new views or physics-aware robotic videos.

\textbf{(4) Matrix Transformation.} This action injection manner is mostly employed in the panoramic video world models in Fig. \ref{fig:action_injection}. IaaW \cite{gui2025image} presents a world exploration interface where users can specify pitch and yaw rotations and adjust the spherical coordinates to simulate view changes. GenEx \cite{lu2024genex} defines an action sequence for sphere rotation sampled from the Unreal Engine or Unity.

\begin{figure*}[t]
\centering
\includegraphics[width=1.0\linewidth]{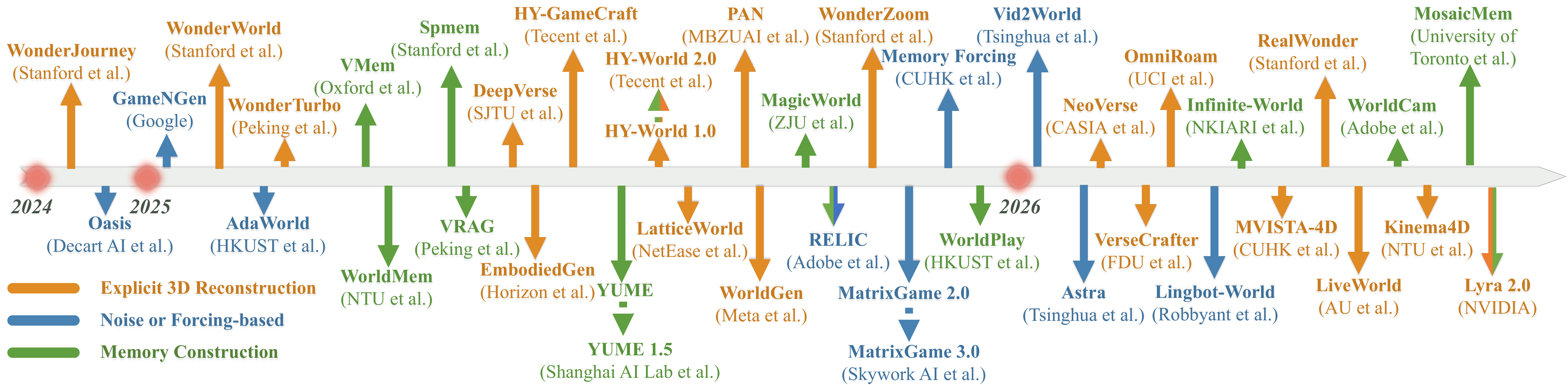}
\vspace{-6mm}
\caption{\textbf{A timeline of works in achieving long-horizon consistency.} We comprehensively review existing methods enabling long-term interactions with Memory Construction methods, Noise or Forcing-based methods, and Explicit 3D Reconstruction methods. Notably, existing approaches mostly use history frames autoregressively, which are not included here.}
\label{fig:long_term}
\vspace{-3mm}
\end{figure*}

\subsubsection{Injection of Text Instructions}
Text instructions offer more flexible and powerful interactive commands combined with pre-trained LLMs \cite{raffel2020exploring}, through which users can personalize their created worlds by freely altering the style, lighting, or layout, or even adding/removing agents. Due to well-established text-to-image diffusion models \cite{cao2025controllable}, such as ControlNet \cite{zhang2023adding}, language instructions can be naturally incorporated as conditioning signals through pretrained text encoders. Existing methods \cite{che2024gamegen, wu2025video, hong2025relic, zhu2026astra,yang2026neoverse} widely leverage T5 \cite{raffel2020exploring} to extract text features and then inject them by cross attention into video tokens. 

However, there are still two challenges to achieving fine-grained action control. On the one hand, fine-grained action injection is harmful to retaining long-duration consistency because the model needs to balance the trade-off between context from prior history frames and immediate action-following response from actions. On the other hand, multi-round controllability poses another challenge to efficiency, where real-time feedback is required for timely interaction iterations. To address these problems, recent efforts are introduced for long-horizon consistency and action-following efficiency in Section \ref{sec:long-term} and Section \ref{sec:response}, respectively.



\section{Long-Horizon Interactions and Memory}\label{sec:long-term}
Predicting long-duration rollouts is a long-standing bottleneck in world models \cite{yu2025survey}, where spatio-temporal consistency is hard to maintain after generating multiple frames. This requirement is further enlarged in the interactive setting with immediate action-following capability. In addition, the commonly-adopted autoregressive paradigm naturally induces accumulated long-term drifts \cite{huang2025self,huang2025memory}, termed as the \textit{Compounding Errors}. To achieve long-duration consistency, various efforts are introduced chronologically as in Fig. \ref{fig:long_term}, including history as context, explicit 3D reconstruction, memory construction, and diffusion training methods such as noise augmentation and Forcing series in DiT \cite{chen2024diffusion}.


\subsection{History Frames as Condition}
\label{sec:History Frames as Condition}
A natural strategy to maintain long-duration consistency is utilizing history frames as context, through which subsequently generated frames can obtain sufficient structural and semantic priors. The concurrent literature \cite{bruce2024genie,wu2024ivideogpt,yang2024learning} mostly adopts an auto-regressive framework, where at least the latest frame would serve as conditional information in the diffusion-based generation model as $\epsilon_{\theta}(o_{t}|o_{t-1}, a_{t-1})$ defined in Section \ref{sec:definition}, where Genie \cite{bruce2024genie} is one of representative works. Follow-up methods further extend the conditional range with more history frames with $\epsilon_{\theta}(o_{t}|o_{\leq t-1}, a_{\leq t-1})$. As a milestone, UniSim \cite{yang2024learning} introduces an overlapping chunk-based generation paradigm, where the last four history frames are concatenated with the current noisy samples in diffusion models. Similarly, later works including DIAMOND \cite{alonso2024diffusion}, AdaWorld \cite{gao2025adaworld}, MineWorld \cite{guo2025mineworld}, HY-GameCraft \cite{li2025hunyuangamecrafthighdynamicinteractivegame}, Matrix-Game \cite{zhang2025matrix,he2025matrix,matrixgame3}, Yume \cite{mao2025yume1,mao2025yume1.5}, PAN \cite{xiang2025pan}, Astra \cite{zhu2026astra}, etc, also reuse multiple past observations, which are concatenated with the next noisy observation channel-wise. To take efficiency into account, most of these methods \cite{hafner2019dream,zhang2025matrix,he2025matrix} design VAE-based token compressor or feature encoders to downsample past videos into latent spaces. However, these chunk-based methods still suffer from limited temporal windows due to fixed-length history frames as conditions. Therefore, an increasing number of studies have adopted the memory mechanism to achieve longer-term coherence.

\subsection{Memory Construction}
\label{sec:Memory Construction}
Inspired by humans, recent efforts have widely explored memory construction in world models \cite{xiaoworldmem}. Memory can act as a 'knowledge base' recalling past observations and remembering how the world state evolves temporally, thereby facilitating counterfactual reasoning ability and revisiting coherence \cite{mao2025yume1,duan2026liveworld}. Additionally, historical observations stored and retrieved from memory can significantly enhance spatio-temporal consistency and empower world modeling with dynamic awareness \cite{wu2026infinite}. Existing memory-dependent methods can be classified into two categories in terms of different stored contents: (1) Memory with implicit 3D consistency, retrieving latent video tokens from history observations. (2) Memory with explicit 3D storage, explicitly keeping 3D geometry with structural coherence.

\subsubsection{Memory with Video Latent Tokens} 
The core concern is about what contents are effectively stored as memory and how to efficiently retrieve the pre-stored memory. WorldMem \cite{xiaoworldmem} first proposes a token-level memory bank that stores all historically created latent tokens augmented by explicit state cues, such as spatial location, viewpoint, and timestamp. The integration of timestamp variable intrinsically captures scene dynamics. Furthermore, a state-aware cross attention is designed to retrieve past observations. VRAG \cite{chen2025learning} develops a memory-retrieval augmented video world model that retrieves related past observations using similarity calculations and injects them into the DiT module. Specifically, position-aware global states, action sequences, and video frames are retrieved together as a memory triplet to condition diffusion-based generation. However, their memory usage increases linearly corresponding to input frames, thereby posing great challenges to real-time long video streaming. To address this issue, RELIC \cite{hong2025relic} divides the memory into an uncompressed KV cache that stores recent observations with sliding windows and also another compressed long-horizon spatial memory cache that keeps all the other history frames with farther distances. The stored information in memory includes historical latent tokens, corresponding actions, and absolute camera pose within the KV cache. HY-WorldPlay \cite{sun2025worldplay} proposes a hybrid memory combining nearest short-term temporal memory and non-adjacent long-term spatial memory. It also re-assigns temporal indices (Temporal Reframing) to deal with extrapolation artifacts caused by growing temporal distances in standard RoPE \cite{li2025cameras}. WorldCam \cite{nam2026worldcam} proposes pose-anchored long-term memory with 3D consistency. HM-World \cite{chen2026out} designs a hybrid memory combining archivists for static backgrounds and vigilant trackers for dynamic subjects to recall out-of-sight dynamics.

\begin{figure}[t]
\centering
\includegraphics[width=1.0\linewidth]{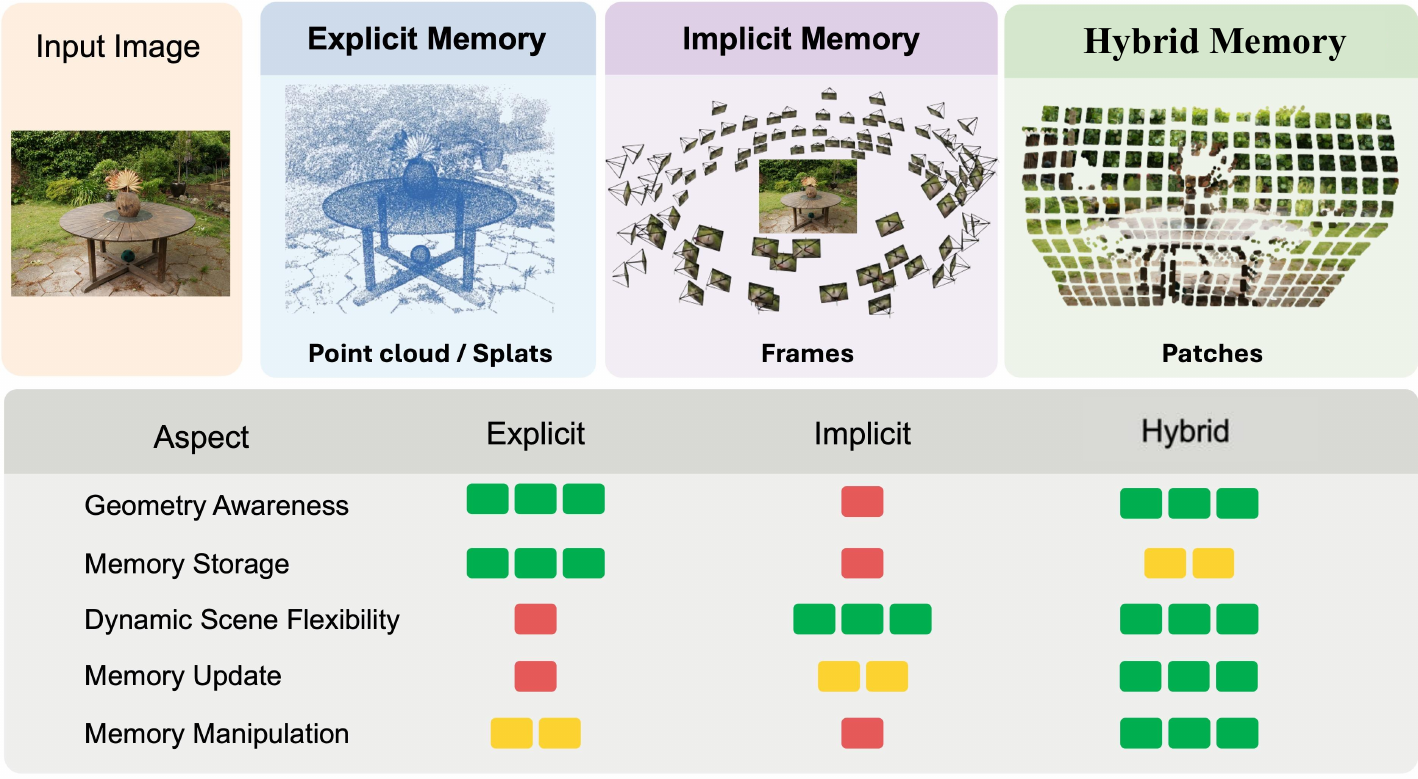}
\vspace{-6mm}
\caption{\textbf{Comparison of different memory constructions.} Explicit memory storing 3DGS \cite{zheng2026versecrafter} or point clouds \cite{li2025magicworld} has better geometry awareness but degraded update feasibility, while implicit memory storing prior video frames has enhanced dynamic handling ability but poor geometric consistency. MosaicMem \cite{yu2026mosaicmem} designs a hybrid memory with mixed advantages from both.}
\label{fig:mem_compare}
\vspace{-3mm}
\end{figure}

\subsubsection{Memory with Explicit 3D Geometric Storage} 
Vmem \cite{li2025vmem} proposes a novel Surfel-indexed view memory which prioritizes past views with the largest overlapping regions with currently generated samples in 3D space. Specifically, memory is read by rendering Surfels with history view indices, where $K$ most frequently-indexing frames are chosen as conditions. Similarly, maintaining an explicit 4D world representation, DeepVerse \cite{chen2025deepverse} retrieves geometrically similar past states prioritizing spatial proximity. Spmem \cite{wu2025video} further subdivides long-term memory into spatial and episodic memory, where the former is represented by an incrementally-updated static point map as intermediate videos generate, while the later one maintains a sparse set of historical reference frames to memorize foreground identities with dynamics. A comparison between explicit and implicit memory is illustrated in Fig. \ref{fig:mem_compare} in different dimensions. Among these, MosaicMem \cite{yu2026mosaicmem} proposes hybrid memories that benefit from explicit and implicit ones. 


\subsection{Explicit 3D Reconstruction}
Without additionally storing history and retrieving memory processes, another representative line of research focuses on directly reconstructing 3D states, such as point cloud, 3D Gaussian Splatting (3DGS), or mesh representations, for consistency. The most well-known literature is Wonder series \cite{yu2025wonderworld,yu2024wonderjourney,ni2025wonderturbo,cao2025wonderzoom}. Given a user-provided image or text as references, these models generate explorable 3D scenes from \textit{anywhere} to \textit{everywhere}. WonderJourney \cite{yu2024wonderjourney} initiates this direction by using an LLM to create text descriptions, guiding the generation of point clouds. Its workflow also contains a training-free VLM to verify the generated scene. WonderWorld \cite{yu2025wonderworld} boosts 3D generation speed with its proposed Fast LAyered Gaussian Surfels (FLAGS) module and also develops a guided depth diffusion model to maintain geometric consistency between frames. Follow-up works \cite{liu2026realwonder,cao2025wonderzoom,li2025magicworld} further advance real-time efficiency \cite{ni2025wonderturbo}, integrate 4D dynamics \cite{chen2025deepverse,zheng2026versecrafter,yang2026neoverse,duan2026liveworld}, or construct simulation-ready dynamics for embodied AI \cite{wang2025embodiedgen,wang2026mvista}. Specifically, WorldGen \cite{wang2025worldgen} introduces a modular 3D world model by combining LLM-driven layout reasoning, procedural generation, and object-specific scene decomposition, which designs a navigation mesh to represent the connectivity of generated blockout. The HY-World series \cite{hy2026hy,team2025hunyuanworld} also belong to this category based on navigable 3DGS or mesh. VDAWorld \cite{o2025vdaworld} resorts to VLMs for grounded 3D representations in simulators and designs a critic prompt to rectify any errors. Some other methods \cite{chen2025physgen3d} focus on physical plausibility with 3D simulation and physics-aware rendering.

\subsection{Noise Augmentation and Forcing-based Training in AR Diffusion}

\begin{figure*}[t]
\centering
\includegraphics[width=1.0\linewidth]{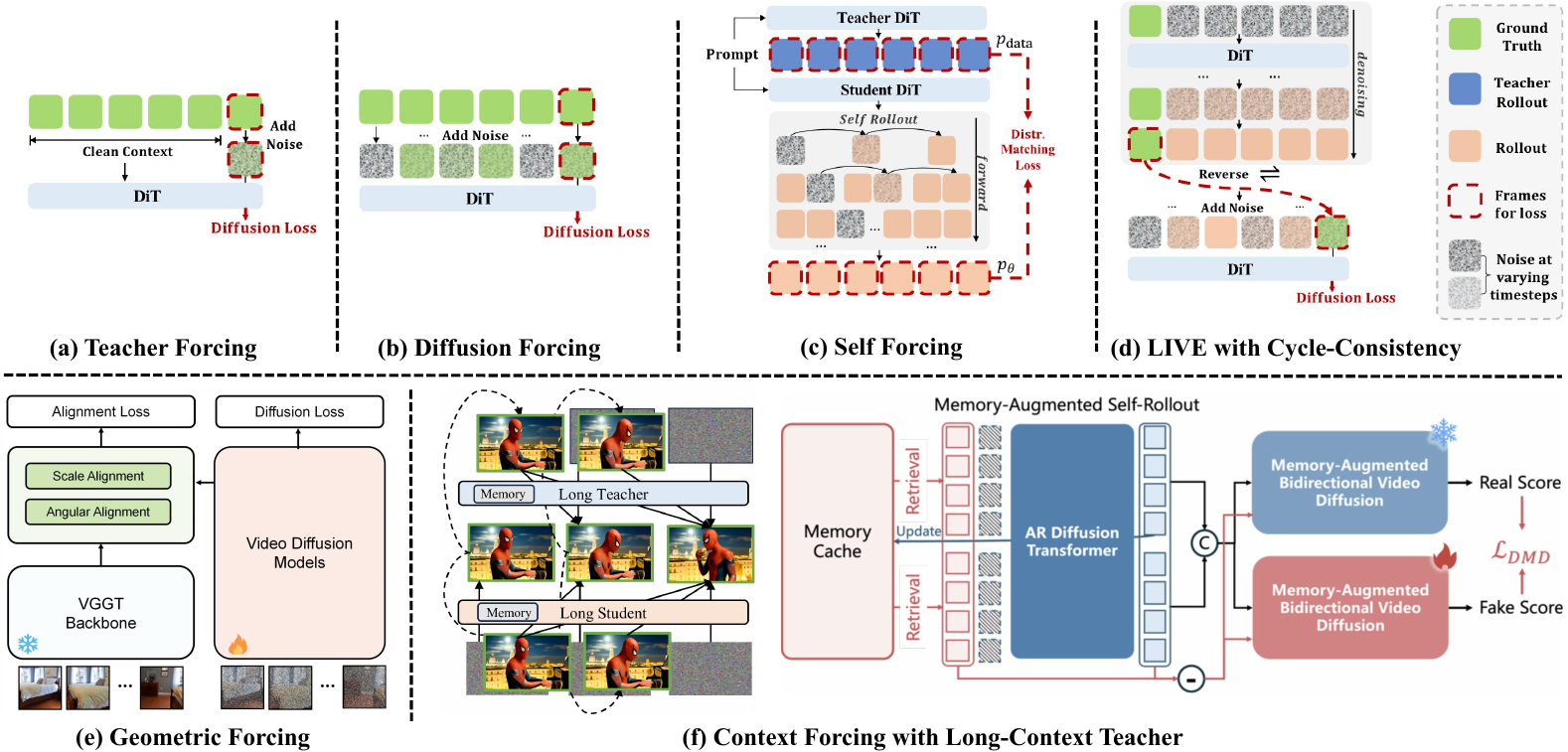}
\vspace{-8mm}
\caption{\textbf{Comparison among various diffusion distillation paradigms.} Recent methods adopt various distillation approaches in DiT to strengthen long-term consistency, mitigate exposure bias, and enable real-time rollout. Teacher Forcing \cite{jin2024pyramidal} uses ground truth context during training, causing train-inference mismatch. Diffusion Forcing \cite{chen2024diffusion} leverages levels of noise but struggles with real rollout errors. Self-Forcing \cite{huang2025self} employs self-rollout distillation with unbounded error accumulation. LIVE \cite{huang2026live} performs forward rollout then reverses recovery with frame-level diffusion loss and the cycle-consistency objection. Geometry Forcing \cite{wu2025geometry} strengthens geometry-aware world states by aligning them with 3D features from a pretrained
geometric foundation model. HY-WorldPlay \cite{sun2025worldplay} and Context Forcing \cite{chen2026context} address the mismatch in training long-context memory-empowered student with short-context and memory-less teacher.}
\label{fig:train_compare}
\vspace{-3mm}
\end{figure*}

Some methods modulate the noise levels during the inference of diffusion process. As a pioneering method, Oasis \cite{oasis2024} adopts a dynamic noising pipeline, where the inference-time noise is adjusted on a defined more-to-less schedule. In this way, the diffusion model can retain high-frequency details in previous frames with enhanced consistency. Noise modulation is also used during training to mitigate \textit{exposure bias}, a phenomenon that autoregressive diffusion models are trained exclusively on the perfect ground-truth context, but
must rely on their own imperfect predictions at inference time \cite{huang2025self}. To address the exposure bias, recent approaches, including AdaWorld\cite{gao2025adaworld}, GameNGen \cite{valevski2025diffusion}, Vid2World \cite{huang2026vidworld}, RELIC \cite{hong2025relic}, Astra \cite{zhu2026astra}, HY-WorldPlay \cite{sun2025worldplay}, Lingbot-World \cite{team2026advancing}, widely deploy Diffusion Forcing \cite{chen2024diffusion}, corrupting historical frames by adding a varying amount of Gaussian noise during training to alleviate long-horizon drift and further improve video quality. RELIC \cite{hong2025relic} mixes both Teacher Forcing \cite{jin2024pyramidal} and Diffusion Forcing strategies, producing optimal initialization for causal distillation. Matrix-Game 2.0 \cite{he2025matrix} introduces Self-Forcing \cite{huang2025self} to further reduce exposure bias and accumulated errors based on self-generated rollouts rather than ground truth.  Geometry Forcing \cite{wu2025geometry} marries video world models with 3D awareness. HY-WorldPlay \cite{sun2025worldplay} ensures long-term consistency by proposing a Context Forcing \cite{chen2026context} mechanism with memory-augmented self-rollouts to mitigate the mismatch in long-context memory-based student with short-context, memory-less teacher. Various Forcing-based methods are compared in Fig. \ref{fig:train_compare}.



There are also other alternative solutions to enable long-duration consistency in interactive world models. Some papers \cite{che2024gamegen,gui2025image} utilize video continuation as a post-processing step. For example, IaaW \cite{gui2025image} investigates a VR-based panoramic world model where users can rotate the created worlds for exploration, and another world continuation model is specially used to extend video segments.

\section{Action-Following Responsiveness for Real-time Interactivity}\label{sec:response}
\subsection{Conflict between Coherence and Action-Following}
It remains an open issue to balance the trade-off between the length of conditional history frames and real-time responsiveness to user actions \cite{zhu2026astra}. As the number of past history observations increases, the long-term consistency of the generated videos can be improved as described in Section \ref{sec:long-term}. However, this weakens the response to actions at the same time because generated future frames depend too much on history while overlooking user action inputs. To mitigate this phenomenon, Astra \cite{zhu2026astra} injects random noise into the conditioning frames to blur their influence on future content, forcing the model to avoid heavy reliance on history and to take instant actions into account. HY-WorldPlay \cite{sun2025worldplay} achieves both long-term consistency and real-time interaction with the proposed Context Forcing module, which develops memory-augmented self-rollout. 

\subsection{Real-Time Rollout with Optimized Efficiency}
For interactive world modeling, it is a crucial requirement of high-speed feedback w.r.t. the controlling actions provided by the users, especially for the real-time gameplay experience \cite{guo2025mineworld}. In autonomous driving and embodied AI scenarios, responsiveness matters not only for visual feedback but also for downstream planning~\cite{zeng2025futuresightdrive}. To achieve this real-time interactivity, various distillation or caching methods have been proposed recently.

\begin{table*}[t]
\centering
\small
\setlength{\tabcolsep}{9pt}
\caption{\textbf{A comprehensive list of commonly-adopted datasets and evaluation benchmarks in Interactive Open-world Exploration, Immersive Game Engines, embodied AI, and Interactive Simulators for Autonomous Driving.} Types I, U, N, G, and A denote Indoor, Urban, Nature, Game, and Artistic scenes, respectively. For the Source column, R indicates Real data and S indicates Synthetic data.}
\label{tab:3d_datasets_RS}
\vspace{-3mm}
\resizebox{\textwidth}{!}{%
\begin{tabular}{l c c c c c c c}
\hline\toprule
\textbf{Dataset} & \textbf{Source} & \textbf{Scale} & \textbf{Scenarios} & \textbf{Multi-view} & \textbf{Camera Control} & \textbf{3D Consistency} & \textbf{Memory} \\
\midrule
\rowcolor{mydark_blue}
\multicolumn{8}{c}{Interactive Open-world Exploration} \\
TC-Bench~\cite{feng2024tc}                  & R          & 150 samples      & U \& N    & \xmark & \xmark & \xmark & \xmark \\
 \rowcolor{myblue}EvalCrafter~\cite{liu2024evalcrafter}               & R          & 700 samples      & I \& U \& N & \xmark & \xmark & \xmark & \xmark \\
FETV~\cite{liu2023fetv}                      & R          & 619 samples      & U \& N  & \xmark & \xmark & \xmark & \xmark \\
 \rowcolor{myblue}VBench~\cite{huang2024vbench}                    & R          & 800 samples    & I \& U \& N & \xmark & \xmark & \xmark & \xmark \\
T2V-CompBench~\cite{li2024t2v}             & R          & 700 samples     & I \& U \& N & \xmark & \xmark & \xmark & \xmark \\
 \rowcolor{myblue}ChronoMagic-Bench~\cite{yuan2024chronomagic}         & R          & 1649 samples      & I \& U \& N  & \xmark & \xmark & \xmark & \cmark \\
 WorldScore~\cite{duan2025worldscore}         & R \& S  & 3000 samples  & I \&U \& N \& A & \xmark & \cmark & \cmark & \cmark \\
\rowcolor{myblue}OmniWorldBench~\cite{omniworldbench2026}   & S  &  1068 samples  & I \& U \& N  & \xmark & \cmark & \cmark & \cmark \\
MIND~\cite{ye2026mind}  & S  & -  & U \& N  & \xmark & \cmark & \cmark & \cmark \\
\rowcolor{myblue}WorldScopeDataset~\cite{ni2025wonderfree}         & R \& S  & 23.4M images  & I \&U \& N \& A & \cmark & \cmark & \cmark & \cmark \\
WorldMark~\cite{xu2026worldmark} & R \& S  & 500 samples  & I \& U \& N \& A \& G & \cmark & \cmark & \cmark & \xmark \\
\rowcolor{myblue}iWorldBench~\cite{fang2026benchmark}  &R \& S  &4900 samples & I \&U \& N&\cmark & \cmark & \cmark & \cmark \\
\midrule
\rowcolor{mydark_blue}
\multicolumn{8}{c}{Immersive Game Engines}\\
Atari 100k \cite{Kaiser2020Model} & S &15M images &G &\xmark &\cmark &\xmark &\xmark\\
\rowcolor{myblue}GameNGen \cite{valevski2025diffusion} & S &70M images &G &\xmark &\cmark & \xmark &\xmark\\
VPT \cite{baker2022video} & S &10M images&G&\xmark &\cmark&\xmark &\xmark\\
\rowcolor{myblue}OGameData \cite{che2024gamegen} & S &4000 hours &G &\xmark &\cmark &\xmark &\xmark\\
Source \cite{feng2025the} & R \& S & 1.95M images&G\&U &\cmark &\cmark &\xmark &\xmark\\
\rowcolor{myblue}GF-Minecraft~\cite{Yu_2025_ICCV} & S & 70 hours &G &\xmark &\cmark &\xmark &\xmark\\
Matrix-Game-MC \cite{zhang2025matrix} & S &3700 hours &G &\xmark &\cmark &\xmark &\xmark\\
\rowcolor{myblue}Matrix-Game 2.0 \cite{he2025matrix} & R \& S &800 hours & G\&N & \xmark &\cmark&\xmark&\xmark\\
Sekai-Real-HQ \cite{mao2025yume1} & R &400 hours & N &\xmark &\cmark&\xmark&\xmark\\
\rowcolor{myblue}RELIC \cite{hong2025relic} & S & 1600 mins &G\&A & \xmark &\cmark &\xmark&\cmark\\
WorldCam-50h \cite{nam2026worldcam} & S &17 hours &G &\xmark &\cmark &\cmark &\cmark\\
\rowcolor{myblue}GameWorld \cite{ouyang2026gameworld} & S & - &G &\xmark &\cmark &\cmark &\cmark\\
WorldMark~\cite{xu2026worldmark} & R \& S  & 500 samples  & I \& U \& N \& A \& G & \cmark & \cmark & \cmark & \xmark \\
\midrule
\rowcolor{mydark_blue}
\multicolumn{8}{c}{Interactive Data Engines in Embodied AI}\\
\rowcolor{myblue}EMMA~\cite{dong2025emma}           & S           & -      & I    & \cmark & \cmark & \xmark & \xmark \\
EmbodieDreamer~\cite{wang2025embodiedreamer}           & S           & -      & I    & \cmark & \cmark & \xmark & \xmark \\
\rowcolor{myblue}MimicDreamer~\cite{li2025mimicdreamer}           & S           & -      & I    & \xmark & \cmark & \xmark & \xmark \\
GigaWorld-0~\cite{team2025gigaworld}                 & S           & -      & I    & \cmark & \cmark & \xmark & \cmark \\
\rowcolor{myblue}Interactive World Simulator~\cite{wang2026interactive}               & S          & -      & I & \xmark & \cmark & \xmark & \cmark \\
MVISTA-4D~\cite{wang2026mvista}               & S           & -      & I & \xmark & \xmark & \cmark & \cmark \\
\rowcolor{myblue}LIBERO~\cite{liu2023libero} & S & - & I & \cmark & \cmark & \cmark & \cmark \\
CALVIN~\cite{zhang2025vlabench} & S & - & I & \cmark & \cmark & \cmark & \cmark \\
\rowcolor{myblue}VLABench~\cite{zhang2025vlabench} & S & - & I & \cmark & \cmark & \cmark & \cmark \\
RoboTwin 2.0~\cite{robotwin} & S & - & I & \cmark & \cmark & \cmark & \cmark \\
\rowcolor{myblue}RoboChallenge~\cite{yakefu2025robochallenge} & R & - & I & \cmark & \cmark & \cmark & \cmark \\

\midrule
\rowcolor{mydark_blue}
\multicolumn{8}{c}{Interactive Simulators for Autonomous Driving} \\
nuScenes~\cite{caesar2020nuscenes}                  & R               & 1.4M images       & U        & \cmark & \xmark & \cmark & \cmark \\
\rowcolor{myblue}INTERACTION~\cite{interaction}            & R               & 40K+ images & U\&N     & \xmark & \xmark & \xmark & \cmark \\
RAD~\cite{gao2025rad}  & S & - & U & \cmark & \cmark & \cmark & \cmark \\
\rowcolor{myblue}ReconDreamer-RL~\cite{ni2025recondreamerrl} & S & -  & U & \cmark & \cmark & \cmark & \cmark \\
Simscale~\cite{tian2025simscale} & S & -  & U & \cmark & \cmark & \cmark & \cmark \\
\rowcolor{myblue}NAVISIM \cite{dauner2024navsim}& R & - & U & \cmark & \cmark & \cmark & \cmark \\
Bench2Drive \cite{jia2024bench2drive}& S & - & U & \cmark & \cmark & \cmark & \cmark \\
\rowcolor{myblue}ACT-Bench \cite{arai2025actbench} & S & - & U & \cmark & \cmark & \cmark & \cmark \\
DrivingGen \cite{zhou2026drivinggen} & S & - & U & \cmark & \cmark & \cmark & \cmark \\
\bottomrule\hline
\end{tabular}%
}
\vspace{-5mm}
\end{table*}

\textbf{Model Distillation in Diffusion Models.} Recent world models commonly achieve real-time response by distilling knowledge from a bi-directional non-causal teacher network to another fewer-step causal network for improved efficiency \cite{huang2025self,nam2026worldcam}. GameNGen \cite{valevski2025diffusion} pushes the generation speed to 50 FPS through distribution distillation \cite{yin2024one} yet with degraded visual quality. HY-GameCraft \cite{li2025hunyuangamecrafthighdynamicinteractivegame} distills the standard diffusion process into a compact 8-step consistency model \cite{song2023consistency}. MotionStream \cite{shin2026motionstream} adopts a Self-Forcing-style distribution matching distillation. To strengthen rollout consistency during distillation, MWM \cite{yan2026mwm} proposes a mobile world model, maintaining action-conditioned consistency by an Inference-Consistent State Distillation module for few-step diffusion. Matrix-Game 3.0 \cite{matrixgame3} designs a multi-segment distillation method with up to 40 FPS. Various Forcing-based distillation methods in autoregressive diffusion models are systematically compared in Fig. \ref{fig:train_compare}.

\textbf{Cache-based Acceleration.}
Caching mechanism is a common strategy in transformer backbones, widely applied in recent LLM training. Yume \cite{mao2025yume1} adopts layer-specific caching policies by reusing intermediate residual features across denoising steps to reduce computational costs. HY-World 1.0 \cite{team2025hunyuanworld} combines both caching and multi-GPU parallelization to enable real-time interaction. Similarly, caching-based acceleration is also deployed in world models including Matrix-Game 2.0 \cite{he2025matrix}, SSM-WM \cite{po2025long}, RELIC \cite{hong2025relic}, etc.

\textbf{Other Efficiency Optimization Approaches.} There are also other inference acceleration methods, such as parallel decoding \cite{team2026advancing,guo2025mineworld}, key-frame reconstruction \cite{yang2026neoverse}, few-step sampling quantization \cite{matrixgame3,hong2025relic}, model pruning \cite{hong2025relic}, etc. We refer readers to one of recent survey papers \cite{he2026video} specially reviewing efficiency-relevant designs in world models.

\section{Benchmarks and Method Comparisons}

\label{sec:evaluate}

\subsection{Datasets and Benchmarks}
In terms of four different application scenarios, we introduce common datasets and benchmarks as in Table \ref{tab:3d_datasets_RS}.

\textbf{Interactive Open-world Exploration.} Several benchmarks have been proposed for evaluating interactive world models in open-world exploration, but early ones primarily focus on visual quality, lacking comprehensive assessment of true interactive capabilities. For instance, TC-Bench~\cite{feng2024tc}, EvalCrafter~\cite{liu2024evalcrafter}, FETV~\cite{liu2023fetv}, VBench~\cite{huang2024vbench}, T2V-CompBench~\cite{sun2025t2v}, and ChronoMagic-Bench~\cite{yuan2024chronomagic} mostly lack evaluation dimensions such as camera controllability, 3D consistency or memory mechanisms. Subsequently, WorldScore~\cite{duan2025worldscore} introduces a unified evaluation framework for world models, incorporating controllability, visual fidelity, and temporal dynamics to enable a more comprehensive assessment. Complementary to this, OmniWorldBench~\cite{omniworldbench2026} proposes 4D interaction-centric evaluation metrics to quantify how actions drive state changes over time, and MIND~\cite{ye2026mind} emphasizes memory consistency and action control. Together, these works expand the evaluation dimensions to achieve interactive world models. However, these benchmarks are designed for model evaluation rather than large-scale training. Inspired by WorldScore, WorldScopeDataset~\cite{ni2025wonderfree} leverages Unreal Engine and various generative models to construct a large-scale interactive dataset for training interactive world models. Nevertheless, its visual fidelity remains somewhat limited compared with real-world captured datasets. More recently, iWorldBench \cite{fang2026benchmark} is specially designed for interactive world model evaluation with mixture of robotic, autonomous driving, 3D reconstruction, and UAVs data sources, considering multiple lighting conditions, trajectory-following ability, and memory.

\textbf{Immersive Game Engines.} Achieving immersive and real-time game engines based on world models poses a great challenge to effective gameplay video collection with action labels. Recent efforts sample human playing resources from 2D Games \cite{liao2025genie}, AAA Games \cite{che2024gamegen,feng2025the}, MineCraft \cite{Yu_2025_ICCV,zhang2025matrix}, or Unreal Engine (UE) \cite{hong2025relic,matrixgame3}, etc. To prevent overfitting on game-only datasets, some works also leverage real-world datasets such as Sekai \cite{mao2025yume1, he2025matrix}. However, concurrent benchmarks commonly ignore the assessment of 3D consistency, multi-view consistency, and memory. Recently, GameWorld \cite{ouyang2026gameworld} proposes a benchmark designed for standardized and verifiable evaluation of MLLMs as generalist game agents.

\textbf{Interactive Data Engines in Embodied AI.} For embodied AI, collecting high-quality interaction data is both expensive and labor-intensive. To mitigate this limitation, several approaches leverage interactive world models to augment training data for embodied manipulation~\cite{dong2025emma,wang2025embodiedreamer,li2025mimicdreamer,lang2026vag,wang2026reconphys}. These methods generalize large-scale datasets by combining real-world or simulated data, and in some cases, transfer human action videos to the robotic domain. In addition, some works \cite{team2025gigaworld,wang2026interactive,wang2026mvista} directly use interactive world models as data engines to expand training data on top of existing datasets and simulations. For example, GigaWorld-0~\cite{team2025gigaworld} integrates controllable video generation with 3D-consistent and physically grounded scene modeling to synthesize large-scale, diverse, and instruction-aligned embodied interaction data. Meanwhile, numerous simulation platforms~\cite{liu2023libero,mees2022calvin,zhang2025vlabench,robotwin} provide high-quality, multi-view datasets that not only support rich interactive operations but also offer strict 3D consistency and long-term Memory. Such datasets are critical for training robust and generalizable interactive world models capable of long-horizon reasoning and complex task execution. Moreover, RoboChallenge~\cite{yakefu2025robochallenge} introduces the first benchmark designed in real-world environments, enabling evaluation of interactive world models under realistic scenarios.

\begin{table}[t]
\centering
\setlength{\tabcolsep}{4pt}
\caption{\textbf{Comparison of 3D interactive world models for open-world exploration~\cite{duan2025worldscore,ni2025wonderfree}.} We compare representative methods in terms of camera controllability, 3D consistency, subjective quality, and automatic quality/alignment metrics. Higher is better for all reported metrics. ``-'' indicates that the corresponding result is not reported in the original paper. Abbreviations: CC = camera controllability, 3DC = 3D consistency, SQ = subjective quality, CLS = CLIP score, CLC = CLIP-based consistency, CLI = CLIP-IQA+, QA = Q-Align, and CLA = CLIP aesthetic score. The first-, second-, and third-best methods in each metric are highlighted using \colorbox{NO1}{orange}, \colorbox{NO2}{yellow}, and \colorbox{NO3}{green}.}
\vspace{-3mm}
\label{tab:combined_3d_eval}
\resizebox{\columnwidth}{!}{%
\small
\begin{tabular}{@{}l*{8}{c}@{}}
\hline\toprule
\textbf{Method}
& \textbf{CC}
& \textbf{3DC}
& \textbf{SQ}
& \textbf{CLS}
& \textbf{CLC}
& \textbf{CLI}
& \textbf{QA}
& \textbf{CLA} \\
\midrule
SceneScape~\cite{fridman2024scenescape}      & 84.99 & 76.54 & 32.75 & - & - & - & - & - \\
Text2Room~\cite{hollein2023text2room}        & \cellcolor{NO1}94.01 & \cellcolor{NO2}88.71 & 36.69 & \cellcolor{NO2}34.58 & 0.835 & 0.543 & 2.359 & 4.912 \\
LucidDreamer~\cite{chung2023luciddreamer}    & 88.93 & \cellcolor{NO1}90.37 & \cellcolor{NO2}58.99 & 31.35 & 0.854 & 0.439 & 2.934 & 5.576 \\
DreamScene360~\cite{zhou2024dreamscene360}   & - & - & - & 30.24 & 0.765 & 0.426 & 2.145 & 4.873 \\
WonderJourney~\cite{yu2024wonderjourney}     & 84.60 & 80.60 & \cellcolor{NO1}66.56 & 28.13 & 0.862 & 0.472 & 3.121 & 5.682 \\
InvisibleStitch~\cite{engstler2024invisible} & \cellcolor{NO2}93.20 & \cellcolor{NO3}88.51 & \cellcolor{NO3}58.50 & - & - & - & - & - \\
WonderWorld~\cite{yu2025wonderworld}         & \cellcolor{NO3}92.98 & 86.87 & 49.81 & \cellcolor{NO3}32.28 & \cellcolor{NO3}0.913 & \cellcolor{NO3}0.560 & \cellcolor{NO3}3.437 & \cellcolor{NO3}6.123 \\
WonderTurbo~\cite{ni2025wonderturbo}         & - & - & - & 32.19 & \cellcolor{NO2}0.922 & \cellcolor{NO2}0.562 & \cellcolor{NO2}3.732 & \cellcolor{NO2}6.173 \\
WonderFree~\cite{ni2025wonderfree}           & - & - & - & \cellcolor{NO1}35.00 & \cellcolor{NO1}0.927 & \cellcolor{NO1}0.563 & \cellcolor{NO1}3.912 & \cellcolor{NO1}6.493 \\
\bottomrule\hline
\end{tabular}%
}
\vspace{-5mm}
\end{table}

\textbf{Interactive Simulators for Autonomous Driving.} In autonomous driving settings, nuScenes~\cite{caesar2020nuscenes} remains useful for open-loop evaluation, but it is limited for evaluating genuine interaction during rollout. NAVSIM~\cite{dauner2024navsim} is designed to test whether world-model-generated or world-model-informed futures improve downstream planning under pseudo-simulation, especially with respect to safety, progress, and comfort. Bench2Drive~\cite{jia2024bench2drive} measures closed-loop driving behavior under diverse interactive scenarios. In contrast, ACT-Bench~\cite{arai2025actbench} directly targets trajectory-conditioned action controllability, making it especially useful for evaluating whether a world model actually follows commanded motion. DrivingGen~\cite{zhou2026drivinggen} further proposes an ego-conditioned track, which provides ego-trajectory instructions and evaluates controllability and trajectory alignment. Furthermore, some world models ~\cite{wang2024drivedreamer,gao2023magicdrive,zhao2025drivedreamer4d,wang2025drivegen3d,ni2025recondreamer,zhao2025recondreamer++} themselves can serve as simulation engines for data generation. Moreover, ReconDreamer-RL~\cite{ni2025recondreamerrl} leverages an interactive world model along with scene reconstruction to synthesize a large-scale dataset, which can be directly used to render multiple viewpoints.

\subsection{Metric Comparison of Representative Methods}

\subsubsection{Interactive Open-World Exploration}
Early methods such as SceneScape~\cite{fridman2024scenescape}, Text2Room~\cite{hollein2023text2room}, LucidDreamer~\cite{chung2023luciddreamer}, and DreamScene360~\cite{zhou2024dreamscene360} generate 3D environments from either text prompts or panoramic imagery. Although these approaches support basic interaction, they often lack full camera controllability, 3D consistency, and have relatively low subjective quality in interactive tasks, as shown in Table~\ref{tab:combined_3d_eval}. Subsequently, methods such as WonderJourney~\cite{yu2024wonderjourney} and InvisibleStitch~\cite{engstler2024invisible} explicitly incorporate interactive mechanisms, improving user-driven exploration and multi-view consistency. WonderWorld~\cite{yu2025wonderworld} further enhances the pipeline by providing more robust multi-view reconstruction and partial alignment metrics. To accelerate the generation process, WonderTurbo~\cite{ni2025wonderturbo} optimizes the entire rendering and reconstruction pipeline, achieving faster generation while maintaining competitive consistency and alignment metrics. Finally, WonderFree~\cite{ni2025wonderfree} introduces a dedicated dataset for interactive tasks, achieving superior performance in CLIP-based metrics, Q-Align, and aesthetic scores.

\subsubsection{Embodied AI}
We review interactive world models in embodied scenarios from two perspectives: robotic manipulation and robot policy learning.

\setlength{\tabcolsep}{4pt}
\begin{table}[t]
\centering
\caption{\textbf{Comparison of video generation and world models in robotic manipulation on WorldArena~\cite{shang2026worldarena}.} 
We report representative models under action following (AF), aesthetic quality (AQ), background consistency (BC), depth accuracy (DA), dynamic degree (DD), flow score (FS), image quality (IQ), and instruction following (IF). Higher is better for all reported metrics.}
\vspace{-3mm}
\label{tab:video_world_core_metrics}
\resizebox{\columnwidth}{!}{%
\small
\begin{tabular}{@{}lcccccccc@{}}
\hline\toprule
\textbf{Method} & \textbf{AF }&\textbf{ AQ} & \textbf{BC} & \textbf{DA} & \textbf{DD} & \textbf{FS} &\textbf{ IQ} & \textbf{IF} \\
\midrule
GigaWorld-1~\cite{team2025gigaworld}                & 0.28 & 41.17 & 86.43 & \cellcolor{NO1}98.44 & 30.52 & 18.64 & \cellcolor{NO3}51.18 & 82.14 \\
Ctrl-World~\cite{guo2025ctrl}                       & 3.90 & 37.05 & 90.30 &  \cellcolor{NO3}93.25 & \cellcolor{NO2}41.82 & \cellcolor{NO1}33.57 & 42.44 & 67.68 \\
Wan2.6~\cite{wan2025wan}                            & \cellcolor{NO1}9.92 & \cellcolor{NO2}44.40 & 84.29 & 75.43 & 33.63 & 22.01 & \cellcolor{NO1}67.36 & \cellcolor{NO2}89.96 \\
ABot\_PhysWorld~\cite{chen2026abot}                 & 2.92 & 38.54 & 88.03 & 83.49 & 34.36 & \cellcolor{NO3}26.55 & 40.78 & \cellcolor{NO3}82.24 \\
CogVideoX~\cite{yang2024cogvideox}                  & 0.00 & 36.30 & 88.38 & 91.09 & 31.47 & 21.77 & 36.23 & 75.58 \\
Veo3.1~\cite{googledeepmind2025veo}                 & \cellcolor{NO2}8.52 & \cellcolor{NO1}48.79 & \cellcolor{NO2}91.67 & 74.27 & 14.66 & 8.26  & \cellcolor{NO2}65.57 & \cellcolor{NO1}97.14 \\
IRASim~\cite{zhu2025irasim}                         & 3.94 & 32.86 & 78.69 & \cellcolor{NO2}93.50 & 21.60 & 11.66 & 41.85 & 65.40 \\
TesserAct~\cite{zhen2025tesseract}                  & 2.36 & \cellcolor{NO3}41.30 & \cellcolor{NO1}92.82 & 73.85 & \cellcolor{NO1}56.49 & 26.28 & 35.45 & 66.80 \\
WoW~\cite{chi2025wow}                               & 4.51 & 38.55 & 90.53 & 73.11 & 32.95 & 23.26 & 46.15 & 56.70 \\
Vidar~\cite{feng2025vidar}                          & \cellcolor{NO3}7.86 & 39.87 & 78.24 & 79.77 & 21.05 & 17.03 & 41.75 & 61.54 \\
Wan2.2~\cite{wan2025wan}                            & 5.39 & 38.32 & 79.37 & 80.57 & 21.46 & 10.91 & 39.53 & 49.84 \\
GigaWorld-0~\cite{team2025gigaworld}                & 7.55 & 40.15 & 88.15 & 63.78 & 39.45 & \cellcolor{NO2}30.89 & 48.81 & 56.14 \\
RoboMaster~\cite{fu2025learning}                    & 3.43 & 37.87 & \cellcolor{NO3}90.80 & 83.77 & \cellcolor{NO3}41.74 & 16.07 & 36.05 & 51.14 \\
Genie Envisioner~\cite{liao2025genie}               & 1.07 & 26.39 & 87.54 & 86.83 & 32.41 & 19.91 & 28.18 & 20.36 \\
\bottomrule\hline
\end{tabular}%
}
\vspace{-5mm}
\end{table}

 \noindent\textbf{Interactive World Models for Robotic Manipulation.} Interactive world models can be used both to evaluate success rates and to generate training data for robot manipulation. We compare recent representative interactive world models for manipulation in Table~\ref{tab:video_world_core_metrics}. Ctrl-World~\cite{guo2025ctrl} proposes a controllable multi-view world model for evaluating and improving generalist robot policies in imagination. GigaWorld-0~\cite{team2025gigaworld} presents a unified world model framework designed as a data engine for VLA learning, with joint synthesis of controllable video and physically grounded 3D interaction data. ABot-PhysWorld~\cite{chen2026abot} introduces a DPO-based post-training framework with decoupled discriminators to suppress unphysical behaviors without sacrificing visual quality, and a parallel context block for precise cross-embodiment action control. IRASim~\cite{zhu2025irasim} emphasizes fine-grained robot-object interaction modeling by introducing a frame-level action-conditioning module to strengthen action-frame alignment. TesserAct~\cite{zhen2025tesseract} is a 4D embodied world model that extends video-based world modeling from 2D appearance prediction to joint RGB-DN generation and 4D scene reconstruction. WoW~\cite{chi2025wow} is an embodied world model that seeks to learn physically grounded world knowledge from large-scale real-world interaction data. RoboMaster~\cite{fu2025learning} is a trajectory-controlled video generation framework for robotic manipulation with phase-wise interaction decomposition. Genie Envisioner~\cite{liao2025genie} unifies world modeling, policy learning, and simulation for robotic manipulation, but its scores in Table~\ref{tab:video_world_core_metrics} indicate that it still lags behind recent embodied world models and strong video-generation baselines on several perceptual and instruction-following metrics.

\noindent\textbf{Interactive World Models for Policy Learning.} Compared to traditional Vision-Language-Action models~(VLA)~\cite{pi_0,pi05,team2026gigabrain,ni2026swiftvla,team2025gigabrain}, directly using an interactive world model as a policy often achieves competitive or stronger performance as in Table \ref{tab:wam_combined}. Video Policy~\cite{hu2024video} treats video generation as a proxy for robot policy learning, showing that modeling future robot behavior in the visual domain can provide strong supervision for action generation. Motus~\cite{motus} presents a unified latent-action world model that jointly models understanding, generation, and action for embodied decision-making. Cosmos Policy~\cite{kim2026cosmos} converts a large pretrained video model into an effective robot policy with minimal adaptation, while its unified action generation, future prediction, and value estimation framework suggests that video models can naturally support both policy execution and model-based planning. LingBot-VA~\cite{lingbotva} treats robot control as a causal world modeling problem by jointly learning policy execution and future frame prediction within a unified autoregressive diffusion framework. GigaWorld-Policy~\cite{ye2026gigaworld} makes a promising step toward action-centered modeling, showing that future visual dynamics can serve as rich auxiliary supervision for policy learning without requiring tight coupling between video prediction and action. This design improves the inference efficiency of world--action models, making them more practical for real-time robotic decision-making. Fast-WAM~\cite{yuan2026fast} questions whether explicit future imagination is necessary for effective world--action modeling in robot control. The comparison in Table~\ref{tab:wam_combined} suggests the promise of world-action modeling on RobotWin 2.0~\cite{robotwin} and Libero~\cite{liu2023libero}.

\begin{table}[t]
\centering
\caption{\textbf{Comparison of world-action models for robotic policy learning on RobotWin 2.0~\cite{robotwin} and Libero~\cite{liu2023libero}.} 
We report task success rates of representative methods. Higher is better for all reported results. ``-'' indicates that the corresponding result is unavailable or not reported. Abbreviations: Rand = randomized setting, Avg = average success rate.}
\vspace{-3mm}
\label{tab:wam_combined}
\resizebox{0.5\textwidth}{!}{
\begin{tabular}{lcccccccc}
\hline\toprule
& \multicolumn{3}{c}{\textbf{RobotWin 2.0}} & \multicolumn{5}{c}{\textbf{Libero}} \\
\cmidrule(r){2-4} \cmidrule(l){5-9}
\multirow[c]{-2}{*}{\textbf{Method}} & \textbf{Clean} & \textbf{Rand} & \textbf{Avg} & \textbf{Spatial} & \textbf{Object} & \textbf{Goal} & \textbf{Long} & \textbf{Avg} \\
\midrule
$\pi_0$~\cite{pi_0}                      & 65.92 & 58.40 & 62.2 & 96.8 & 98.8 & 95.8 & 85.2 & 94.2 \\
$\pi_{0.5}$~\cite{pi05}                  & 82.74 & 76.76 & 79.8 & \cellcolor{NO1}98.8 & 98.2 & \cellcolor{NO2}98.0 & 92.4 & 96.9 \\
Motus~\cite{motus}                       & \cellcolor{NO3}88.66 & 87.02 & \cellcolor{NO3}87.8 & 96.8 & \cellcolor{NO2}99.8 & 96.6 & \cellcolor{NO2}97.6 & \cellcolor{NO2}97.7 \\
LingBot-VA~\cite{lingbotva}              & \cellcolor{NO1}92.90 & \cellcolor{NO2}91.50 & \cellcolor{NO1}92.2 & \cellcolor{NO2}98.5 & \cellcolor{NO3}99.6 & \cellcolor{NO3}97.2 & \cellcolor{NO1}98.5 & \cellcolor{NO1}98.5 \\
GigaWorld-Policy~\cite{ye2026gigaworld}  & 86.21 & \cellcolor{NO3}87.34 & 86.8 & - & - & - & - & - \\
Fast-WAM~\cite{yuan2026fast}             & \cellcolor{NO2}91.88 & \cellcolor{NO1}91.78 & \cellcolor{NO2}91.8 & \cellcolor{NO3}98.2 & \cellcolor{NO1}100.0 & 97.0 & \cellcolor{NO3}95.2 & \cellcolor{NO3}97.6 \\
Video Policy~\cite{hu2024video}          & - & - & - & - & - & - & 94.0 & - \\
Cosmos Policy~\cite{kim2026cosmos}       & - & - & - & 98.1 & \cellcolor{NO1}100.0 & \cellcolor{NO1}98.2 & \cellcolor{NO2}97.6 & \cellcolor{NO1}98.5 \\
\bottomrule\hline
\end{tabular}
}
\vspace{-5mm}
\end{table}

\subsubsection{Immersive Game Engines}
Recent game-oriented world models~\cite{mao2025yume1,sun2025worldplay,team2026advancing,he2025matrix,hong2025relic,li2025hunyuangamecrafthighdynamicinteractivegame} have received increasing research focus, where visual fidelity, controllable dynamics, long-horizon consistency, and real-time responsiveness are optimized jointly.

We list the capability-level comparison among recent game engines in Table \ref{tab:game_engine}, in dimensions of data source, action space, resolution, generation speed, interaction duration, memory, and model size. From the table, recent methods achieve more flexible action controls, higher resolution, faster inference, longer generated contents, and memory enhancement with smaller model size. Among these, Yume series~\cite{mao2025yume1.5,mao2025yume1} proposes an efficient and text-controllable interactive world generation framework that improves long-horizon context modeling and streaming inference for real-time exploration. Matrix-Game series~\cite{he2025matrix,matrixgame3,zhang2025matrix} frames game-world generation as a real-time streaming problem with explicit action control, combining scalable game-data construction with efficient autoregressive diffusion. HY-GameCraft~\cite{li2025hunyuangamecrafthighdynamicinteractivegame} proposes a 3D interactive world modeling framework that improves action controllability and long-term geometric consistency through hybrid history-conditioned training and model distillation, showing strong imaging quality and competitive aesthetic quality. GameGen-X~\cite{che2024gamegen} is an open-world game video generation and control framework that combines large-scale gameplay data and a modular instruction-tuning design to enable high-quality and controllable gameplay simulation. RELIC~\cite{hong2025relic} further highlights real-time long-horizon generation with consistent spatial memory and precise user control, making memory-aware interaction a central design. WorldCam \cite{nam2026worldcam} designs a 3D world model supporting 6-DoF player actions as input.

\begin{table}[t]
\centering

\caption{\textbf{Comparison of interactive game engines.} Notably, the reported speed largely depends on the model size, GPU hardware, and the
degree of parallelism. We retrieve these numbers from the original paper. T, R, and E are translation-, rotation-, and other interactive related actions, respectively.
}
\vspace{-3mm}
\resizebox{\columnwidth}{!}{%
\begin{tabular}{lcccccc}
\hline\toprule
\rowcolor{mydark_blue}\textbf{Method} & \textbf{Action Space}  & \textbf{Resolution} & \textbf{Speed} & \textbf{Duration} & \textbf{Memory} & \textbf{Size} \\
\midrule

\rowcolor{myblue}Genie~\cite{bruce2024genie}  & Latent &360p&--- &2 s&\xmark&11B\\
DIAMOND~\cite{alonso2024diffusion}&---&280 $\times$ 150 & 15 FPS&Infinite &\xmark&13M\\
\rowcolor{myblue}GameNGen~\cite{valevski2025diffusion}  &---& 240p&20 FPS&Infinite&\xmark&---\\
Oasis~\cite{oasis2024} &4T4R1E&640 $\times$ 360&20 FPS&Infinite&\xmark&500M\\
\rowcolor{myblue}GameGen-X~\cite{che2024gamegen}&4T1E&320p &20 FPS&4-16 s&\xmark&---\\
The Matrix~\cite{feng2025the} &4T4R&720p&16 FPS&Infinite&\xmark&2.7B\\
\rowcolor{myblue}Genie 2~\cite{parkerholder2024genie2} &5T4R2E&720p&---&10-20 s&\cmark & ---\\
HY-GameCraft~\cite{li2025hunyuangamecrafthighdynamicinteractivegame} &4T4R &720P&24 FPS&1 min&\xmark&13B\\
\rowcolor{myblue}GameFactory~\cite{Yu_2025_ICCV} &4T4R &640 $\times$ 360& ---& --- &\xmark&---\\
Matrix-Game~\cite{zhang2025matrix} &4T4R &720p&---&---&\xmark&17B\\
\rowcolor{myblue}Yume \cite{mao2025yume1}  &4T4R&544$\times$960&16 FPS &20 s&\xmark&14B\\
Yan~\cite{ye2025yan}  &7T2R &1080p&60 FPS&Infinite &\xmark&---\\
\rowcolor{myblue}Matrix-Game 2.0 \cite{he2025matrix} &4T &352$\times$640&25 FPS&1 min&\xmark&1.3B\\
Genie 3 \cite{genie3}&5T4R1E&704$\times$ 1280&24 FPS&1 min&\cmark&---\\

\rowcolor{myblue}RELIC~\cite{hong2025relic}  &6T6R &480 $\times$ 832 & 16 FPS & 20 s & \cmark &14B\\
Yume-1.5~\cite{mao2025yume1.5}  &4T4R&704$\times$1280& 16 FPS & ---&\cmark&5B\\
\rowcolor{myblue}Matrix-Game 3.0 \cite{matrixgame3} &4T&720p&40 FPS &---&\cmark&5B\\
WorldCam \cite{nam2026worldcam}  &6-DoF&480 $\times$ 832 &20 FPS & 10 s & \cmark &---\\
\bottomrule\hline
\end{tabular}%
}
\label{tab:game_engine}
\vspace{-5mm}
\end{table}

\subsubsection{Interactive Simulators for Autonomous Driving}
In autonomous driving scenarios, an interactive driving world model should expose an explicit interface that can steer future rollout at inference time, such as ego actions, future trajectories, visual reasoning, or structured scene controls. Under this criterion, we focus on methods that support controllable future generation or tightly coupled generation--planning, rather than latent-only planning methods without an explicit rollout interface. Detailed metrics are listed in Table \ref{tab:drivinggen_video_gen} and Table \ref{tab:nuscenes_video_gen}, corresponding to the results on DrivingGen \cite{zhou2026drivinggen} and NuScenes \cite{caesar2020nuscenes} datasets, respectively.

\noindent\textbf{Controllable Simulators Conditioned by Scene-Level Controls.}
The first line of work involves interaction as scene-level controllable future generation.
DriveGAN~\cite{kim2021drivegan} is an early representative method that learns a differentiable neural simulator from video-action sequences while exposing controls such as weather and object placement.
DriveDreamer~\cite{wang2024drivedreamer} constructs controllable world modeling by learning real driving scenarios with structured traffic constraints.
DriveDreamer-2~\cite{zhao2025drivedreamer2} further enriches the interaction interface by translating user queries into trajectories and HD Maps, enabling generation of customized and long-tail driving events in a user-friendly manner.
These works frame interactive driving world models as scene-level controllable simulators, where the main goal is to synthesize editable driving scenarios under structured external controls rather than to explicitly model the conversion between video and trajectory.

\begin{table*}[t]
\centering
\caption{\textbf{Comparison of generative video or world models in autonomous driving proposed by DrivingGen \cite{zhou2026drivinggen}.}
Metrics are grouped into four aspects: distribution, quality, temporal consistency, and trajectory alignment. FVD (Fr\'echet Video Distance) and FTD (Fr\'echet Trajectory Distance) measure distributional similarity in video space and trajectory embedding space, respectively. ADE (Average Displacement Error) measures the mean pointwise distance between the generated and conditioning trajectories, while DTW (Dynamic Time Warping) measures their overall path discrepancy under non-linear temporal alignment.  }
\vspace{-3mm}
\label{tab:drivinggen_video_gen}

\resizebox{\textwidth}{!}{%
\small
\setlength{\tabcolsep}{3pt}
\begin{tabular}{@{}l|*{1}{c}|*{12}{c}@{}}
\hline\toprule
\multirow{3}{*}{\makecell[l]{Methods}} &
\multirow{3}{*}{\makecell[c]{Size}}
& \multicolumn{2}{c}{Distribution} &
\multicolumn{3}{c}{Quality} &
\multicolumn{4}{c}{Temporal Consistency} &
\multicolumn{2}{c}{Trajectory Alignment} \\
\cmidrule(lr){3-4}
\cmidrule(lr){5-7}
\cmidrule(lr){8-11}
\cmidrule(lr){12-13}
& & \rotcell{FVD $\downarrow$ } & \rotcell{FTD $\downarrow$} &
\rotcell{Subjective\\Quality $\uparrow$} &
\rotcell{Objective\\Quality $\uparrow$} &
\rotcell{Trajectory\\Quality $\uparrow$} &
\rotcell{Video\\Consist. $\uparrow$} &
\rotcell{Agent\\Consist. $\uparrow$} &
\rotcell{Agent\\Missing $\uparrow$} &
\rotcell{Trajectory\\Consist. $\uparrow$} &
\rotcell{ADE $\downarrow$} &
\rotcell{DTW $\downarrow$}  \\
\midrule

Vista~\cite{gao2024vista} & 2.5B & \cellcolor{NO1}{392.8} & \cellcolor{NO1}{27.33} &
0.4146 & \cellcolor{NO3}{0.8198} & \cellcolor{NO1}{0.6047} & \cellcolor{NO3}0.8741 & \cellcolor{NO3}{0.6417} & \cellcolor{NO3}0.8676 & \cellcolor{NO1}{0.4366} & \cellcolor{NO1}{19.70} & \cellcolor{NO1}{1216} \\

UniFuture~\cite{liang2025unifuture} & 3.0B & {654.6} & \cellcolor{NO3}{37.17} &
0.4006 & \cellcolor{NO1}{0.9685} & \cellcolor{NO2}{0.5353} & \cellcolor{NO2}{0.8759} & 0.5525 & \cellcolor{NO2}{0.8759} & \cellcolor{NO2}{0.4165} & \cellcolor{NO2}{20.21} &\cellcolor{NO2}{1352} \\

VaViM~\cite{bartoccioni2025vavim} & 1.2B & 1222 & 103.6 &
\cellcolor{NO1}{0.4910} & \cellcolor{NO2}{0.8694} & 0.1936 & \cellcolor{NO1}{0.9428} & \cellcolor{NO1}{0.8290} & \cellcolor{NO1}{0.9725} & 0.0984 & 41.92 & 3863 \\

DrivingDojo~\cite{wang2024drivingdojo} & 2.3B & \cellcolor{NO3}586.5 & \cellcolor{NO2}{35.73} &
\cellcolor{NO3}{0.4264} & \cellcolor{NO3}{0.8198} & 0.4131 & {0.8419} & \cellcolor{NO2}{0.6940} & {0.8439} & 0.2776 & \cellcolor{NO3}{25.50} & 2142 \\

GEM~\cite{hassan2024gem} & 2.1B & \cellcolor{NO2}{579.9} & 97.70 &
\cellcolor{NO2}{0.4484} & 0.8018 & \cellcolor{NO3}{0.5085} & 0.7886 & 0.6180 & 0.7463 & \cellcolor{NO3}{0.2983} & 25.73 & \cellcolor{NO3}{1982} \\
\bottomrule\hline
\end{tabular}%
}
\vspace{-4mm}
\end{table*}

\begin{table}[t]
\centering
\caption{\textbf{Comparison of video generation models and world models on the NuScenes validation set \cite{caesar2020nuscenes}.}}
\label{tab:nuscenes_video_gen}
\vspace{-3mm}
\small
\setlength{\tabcolsep}{4pt}
\renewcommand{\arraystretch}{1.05}
\resizebox{\columnwidth}{!}{%
\begin{tabular}{lccccc}
\hline\toprule
\textbf{Metric} 
& DriveGAN~\cite{kim2021drivegan}
& DriveDreamer~\cite{wang2024drivedreamer}
& DrivingGPT~\cite{chen2025drivinggpt}
& DrivingWorld~\cite{hu2024drivingworld}
& Vista~\cite{gao2024vista} \\
\midrule
FID $\downarrow$ 
& 73.4 
& 52.6 
& 12.8 
& 7.4 
& \cellcolor{NO3}6.9 
\\

FVD $\downarrow$ 
& 502.3 
& 452.0 
& 142.6 
& 90.9 
& 89.4 \\
\midrule
\textbf{Metric} & Epona~\cite{zhang2025epona}
& DriveDreamer-2~\cite{zhao2025drivedreamer2}
& MagicDrive-V2~\cite{gao2025magicdrive}
& OmniNWM~\cite{li2025omninwm}
& DriveLaW~\cite{xia2025drivelaw}\\
\midrule
FID $\downarrow$ & 7.5
& 11.2
& 20.9
& \cellcolor{NO2}5.5
& \cellcolor{NO1}{4.6} \\

FVD $\downarrow$ & 82.8
& \cellcolor{NO2}55.7
& 94.8
& \cellcolor{NO1}{23.6}
& \cellcolor{NO3}81.3 \\
 
\bottomrule\hline
\end{tabular}%
}
\vspace{-5mm}
\end{table}

\noindent\textbf{Joint Video-Action Generation for Tighter Planning Coupling.}
The second line pushes the interaction deeper into the model formulation itself by representing observation and control signals within one generative process.
DrivingGPT~\cite{chen2025drivinggpt} introduces a multimodal driving language with interleaved image and action tokens, so that world modeling and planning are learned jointly through autoregressive next-token prediction.
Vega~\cite{zuo2026vega} extends this direction by treating natural language instructions as an explicit control interface for personalized driving.
Compared to classical controllable simulators, these methods internalize actions, instructions, or planning signals into the generative token space, enabling tighter coupling between visual imagination and policy learning.

\noindent\textbf{Factorized Video--Trajectory Rollout Paradigms.}
The third line explicitly studies how visual imagination and motion planning are connected through a factorized rollout pipeline.
In trajectory-to-video formulations, an ego trajectory, action sequence, or policy rollout is first specified, and the model synthesizes future driving videos that should remain aligned with this motion condition.
DrivingWorld~\cite{hu2024drivingworld} develops a GPT-style world model for controllable long-horizon future video generation, emphasizing temporal coherence and rollout stability.
Epona~\cite{zhang2025epona} further supports flexible-length, trajectory-conditioned video--trajectory rollout with an autoregressive diffusion formulation.
In this context, Table~\ref{tab:drivinggen_video_gen} compares representative ego-conditioned generative video/world models on DrivingGen~\cite{zhou2026drivinggen}, including Vista~\cite{gao2024vista}, UniFuture~\cite{liang2025unifuture}, VaViM~\cite{bartoccioni2025vavim}, DrivingDojo~\cite{wang2024drivingdojo}, and GEM~\cite{hassan2024gem}.
These methods are evaluated by video realism, temporal consistency, and trajectory alignment, making them closest to the trajectory-conditioned rollout paradigm.
Conversely, in video-to-planning formulations, generated futures or video-generator latents serve as intermediate imagination for downstream trajectory prediction or planning.
DriveLaW~\cite{xia2025drivelaw}, for example, improves imagination--planning consistency by feeding latent representations from the video generator to the planner.
This line therefore focuses on the ordering, conversion, and consistency between video and trajectory generation, complementing scene-level controllable simulators and fully unified video-action models.




\section{Challenges and Future Directions}
\label{sec:future}

Despite rapid progress, several fundamental challenges remain unresolved towards next-generation interactive world modeling, especially when these models are expected to support reliable decision-making and safety-critical environments \cite{ding2025understanding,guan2024world,lecun2022path}. 

\subsection{Large-Scale Action-Labeled Data Acquisition}
 \textbf{Existing Challenges.} A major bottleneck lies in the acquisition of large-scale action-conditioned data. Existing interactive world models often rely on substantial amounts of paired observations and actions, whose collection is expensive, domain-specific, and difficult to scale. This issue becomes even more severe in real-world settings such as autonomous driving and robotics, where accurate action annotations, state transitions, and safety-critical events are all costly to obtain \cite{guan2024world}. Although recent studies have started to automatically learn from unlabeled videos, the discovered actions are often only weakly grounded in true control semantics and can be entangled with scene appearance, camera motion, or dataset bias \cite{baker2022video,gao2025adaworld,wu2024ivideogpt}. Thus, current models still face a gap between scalable pre-training and controllable action-grounded world simulation.

\noindent \textbf{Potential Solutions.} A promising direction is to combine self-supervised latent action with stronger structural priors, such as kinematic constraints, object-centric interaction cues, and geometric consistency. Instead of treating action purely as an implicit latent code, future work could learn representations that are both scalable and semantically interpretable, so that they can be incorporated into large-scale pre-training while remaining effective across downstream tasks, scenarios, embodiments, and sensors \cite{gao2025adaworld,zhu2025aether,wu2025geometry}.

\subsection{Counterfactual Reasoning \& Real-World Alignment}
 \textbf{Existing Challenges.} Current interactive world models have demonstrated increasing controllability through visual or language interfaces, but still rely primarily on statistical correlations rather than explicit counterfactual mechanisms \cite{wu2024ivideogpt,wang2024drivewm,chen2025drivinggpt,zuo2026vega}. As a result, they may generate visually plausible futures while failing to preserve correct cause-and-effect relationships between actions, agent responses, and environment transitions. This limitation becomes especially problematic in safety-critical domains such as autonomous driving, where visually coherent generation does not necessarily imply reliable closed-loop behavior \cite{guan2024world,zhou2026drivinggen}. More broadly, the gap between synthetic controllability and real-world validity remains an obstacle to the deployment of interactive world models in actual decision-making scenes.

\noindent \textbf{Potential Solutions.} Looking forward, an important direction is to endow world models with stronger causal grounding and real-world supervision. This may involve integrating object-centric state abstractions, intervention-based data construction, and structured dynamics priors that explicitly capture how actions alter future world states. In addition, grounding learned simulators in real-world spatial layouts, prior rules, and reinforcement-learning validation may improve robustness under corner cases and distribution shifts \cite{zhu2025aether,wu2025geometry,dai2025fantasyworld,seo2026grounding}. A representative work is SWM, grounding world simulation in Seoul \cite{seo2026grounding}. 


\subsection{Extreme Long-Horizon Consistency}
\textbf{Existing Challenges.} Maintaining coherent world evolution over extremely long horizons remains one of the most critical open problems. Although recent methods can produce minute-level long-term interactive rollouts with promising fidelity \cite{hong2025relic}, it is still non-trivial to extend them to hour-level or even longer ones. Furthermore, current methods often suffer from temporal drift, structural inconsistency, object disappearance, and unstable latent world states when generation extends over longer horizons \cite{chen2026context,yang2026stableworld,zhan2026perpetualwonder}. This challenge is amplified in interactive settings, where the model must not only preserve scene continuity, but also remain responsive to newly injected user controls over multiple rounds of interactions \cite{feng2025the,li2025vmem}.

\noindent \textbf{Potential Solutions.} Future progress will likely require more explicit memory and state-tracking mechanisms. Rather than relying solely on implicit hidden states, next-generation models may need hierarchical memory designs that separately maintain short-term motion details and long-term scene dependencies \cite{wu2025corgi,huang2025memory,li2025vmem,yu2026mosaicmem,chen2026out}. Another promising direction is to combine latent video generation with explicit geometric indexing or global world-state recall, so that the model can better preserve infinite memory and closed-loop recognition during extended rollouts \cite{wu2025geometry,dai2025fantasyworld,nam2026worldcam}. Meanwhile, techniques that reduce train--test mismatch and improve autoregressive stability also play an important role in long-horizon interaction \cite{huang2025self,chen2026context}. 

\subsection{Physical Awareness During Interactions}
 \textbf{Existing Challenges.} Physical awareness is crucial to real-world deployments, especially in safety-required scenarios like self-driving planning \cite{chen2025drivinggpt}, robotic manipulation \cite{lingbotva}, etc. Generated worlds should learn how objects move following Newton's Laws of Motion, how to distinguish the manipulation differences in rigid or deformable objects, and how to generate force conditioned videos. However, it is still challenging to enable physics-grounded imaginations in interactive world models. Most of recent methods \cite{yu2025wonderworld,hong2025relic,wang2024drivedreamer} adopt a data-driven paradigm for video generation, without explicitly modeling physical information. Furthermore, classical video diffusion models mainly target at the denoising objective, potentially imitating physical movements but lacking real physical awareness. 

\noindent \textbf{Potential Solutions.} Some recent methods \cite{liu2026realwonder,zhan2026perpetualwonder,chen2025physgen3d} resort to physical simulators to estimate the material of objects, converting physical action conditions into visual signals such as optical flow \cite{liu2026realwonder} or binding physics particles to strengthen visual primitives \cite{zhan2026perpetualwonder}. However, simulation results may be affected by the quality of explicit geometry constraints. Future approaches could introduce large-scale foundation models with dense structural priors \cite{wang2025vggt} to improve simulation effectiveness. Furthermore, insights from the recent physics-informed video generation domain can be transferred, such as the tight coupling of physical simulator in the generation loop \cite{foo2026physical} or the design of post-verification or detection methods of physical consistency \cite{zhou2026toward} to prevent models from simply imitating physics-similar movements. As a representative method, VDAWorld \cite{o2025vdaworld} abstracts scene physics with VLMs and generates future predictions by interactively executing the simulator, where a Critic Prompt module is used to correct errors.

\subsection{Accessible Interfaces with Direct Manipulation}
 \textbf{Existing Challenges.} A key to realizing interactive world models is to bestow them with interaction techniques that provide users with high-fidelity control, in-depth interpretability, and a high sense of agency and locus of control. The principle of \emph{direct manipulation}~\cite{shneiderman1983direct} is credited with explaining why graphical user interfaces have been so successful, in large part due to its ability to achieve interaction with the above three qualities for deterministic desktop interfaces. The principle states that user interactions should provide three properties: (1) objects and actions of interest to the user should be made visible with meaningful visual metaphors; (2) actions should be rapid, reversible and provide immediate feedback; and (3) replacing the need to specify actions, for instance by typing commands or prompts, with the ability to directly press or select actions that are readily visible to the user. Prior work has demonstrated that partially implementing this principle can induce more immersive user experience~\cite{masson2024directgpt}.

\noindent \textbf{Potential Solutions.} The key to realize the principle of direct manipulation for interactive world models will be to link concepts that are meaningful to users, such as visual objects, textures, behavior, and so on, to representations within the world model that can be both visualized in meaningful ways to users and ensure user interaction enacts intended changes in the underpinning model. Another possible solution is to generate more accessible and versatile user interfaces for manipulation depending on multi-modal context or particular user queries~\cite{vaithilingam2024dynavis}.

\section{Conclusion}
In this article, we systematically reviewed recent advances in interactive world modeling. We revisit recent research trends in application scenarios, world states, and interaction modalities. Three crucial technique bottlenecks are introduced, including user controllability from actions; long-horizon interaction and memory; and action-following responsiveness for real-time interactivity. To facilitate comparison, we comprehensively list existing benchmarks across four domains: game engines, autonomous driving, open-world exploration, and embodied AI. Finally, the last section looks forward to future potential research directions. Please refer to the curated lists with frequent updates in \url{https://github.com/liujiuming123/Awesome-Interactive-World-Model}.

{\bibliographystyle{IEEEtran} 
\bibliography{main} }

\end{document}